\documentclass[journal]{IEEEtran}
\usepackage{amsmath,amsfonts}
\usepackage{algorithmic}
\usepackage{array}
\usepackage[caption=false,font=normalsize,labelfont=sf,textfont=sf]{subfig}
\usepackage{textcomp}
\usepackage{stfloats}
\usepackage{url}
\usepackage{verbatim}
\def\BibTeX{{\rm B\kern-.05em{\sc i\kern-.025em b}\kern-.08em
    T\kern-.1667em\lower.7ex\hbox{E}\kern-.125emX}}
\usepackage{balance}

\usepackage{xspace}
\usepackage{microtype} 
\usepackage{booktabs}  
\usepackage{color}
\usepackage[T1]{fontenc}

\usepackage{tabularx}

\usepackage{adjustbox} 
\usepackage{array}     
\newcolumntype{L}[1]{>{\raggedright\let\newline\\\arraybackslash\hspace{0pt}}m{#1}}
\newcolumntype{C}[1]{>{\centering\let\newline\\\arraybackslash\hspace{0pt}}m{#1}}
\newcolumntype{R}[1]{>{\raggedleft\let\newline\\\arraybackslash\hspace{0pt}}m{#1}}

\makeatletter
\renewenvironment{cases}[1][\lbrace]{%
	\def\@ldelim{#1}
	\matrix@check\cases\env@cases
}{%
	\endarray\right.%
}
\patchcmd{\env@cases}{\lbrace}{\@ldelim}{}{}
\makeatother

\DeclareUnicodeCharacter{0301}{\'{e}}
\DeclareUnicodeCharacter{0308}{\"{i}}

\newcommand{\ops}{optimization problems\xspace}

\newcommand{\cops}{continuous \ops}

\newcommand{\AACT}{AACT\xspace}

\newcommand{\funANV}{functional ANOVA\xspace}
\newcommand{\fANV}{f-ANOVA\xspace}

\newcommand{\Tbl}{Table\xspace}

\newcommand{\Sect}{Section\xspace}

\newcommand{\Fig}{Figure\xspace}
\newcommand{\Figs}{Figures\xspace}

\newcommand{\PSOX}{PSO\textsl{-X}\xspace}
\newcommand{\metafor}{\textsf{METAFO}$\mathbb{R}$\xspace}
\newcommand{\irace}{\texttt{irace}\xspace}
\newcommand{\paramils}{\texttt{ParamILS}\xspace}
\newcommand{\smac}{\texttt{SMAC}\xspace}

\newcommand{\modCMAES}{modCMAES\xspace}

\newcommand{\modDE}{modDE\xspace}

\newcommand{\pso}{particle swarm optimization\xspace}
\newcommand{\PSO}{PSO\xspace}

\newcommand{\es}{evolution strategy\xspace}

\newcommand{\ESs}{ESs\xspace}

\newcommand{\de}{differential evolution\xspace}
\newcommand{\DE}{DE\xspace}

\newcommand{\Topo}{{\textsf{topology}}\xspace}
\newcommand{\Moi}{{\textsf{modelOfInfluence}}\xspace}
\newcommand{\Pop}{{\textsf{population}}\xspace}

\newcommand{\OmgCS}{{\textsf{omega1CS}}\xspace}
\newcommand{\rndMtx}{{\textsf{randomMatrix}}\xspace}
\newcommand{\AcCof}{{\textsf{accelCoeffCS}}\xspace}
\newcommand{\DNPP}{{\textsf{DNPP}}\xspace}
\newcommand{\PertRnd}{{\textsf{Pert$_{\mathrm{\textsf{rand}}}$}}\xspace}
\newcommand{\PertInf}{{\textsf{Pert$_{\mathrm{\textsf{info}}}$}}\xspace}
\newcommand{\none}{{\textsf{none}}\xspace}
\newcommand{\PopConst}{{\textsf{Pop-constant}}\xspace}
\newcommand{\PopIncre}{{\textsf{Pop-incremental}}\xspace}
\newcommand{\PopTV}{{\textsf{Pop-time-varying}}\xspace}
\newcommand{\InitRandom}{{\textsf{Init-random}}\xspace}
\newcommand{\InitHorizontal}{{\textsf{Init-horizontal}}\xspace}
\newcommand{\DNPPRect}{{\textsf{DNPP-rectangular}}\xspace}
\newcommand{\DNPPSphe}{{\textsf{DNPP-spherical}}\xspace}
\newcommand{\DNPPAddStoch}{{\textsf{DNPP-additive stochastic}}\xspace}

\newcommand{\OperatorN}{{\textsf{DNPP-Gaussian}}\xspace}

\newcommand{\PertGau}{{\textsf{\PertInf-Gaussian}}\xspace}
\newcommand{\PertLev}{{\textsf{\PertInf-L{\'e}vy}}\xspace}
\newcommand{\PertUni}{{\textsf{\PertInf-uniform}}\xspace}
\newcommand{\PertRect}{{\textsf{\PertRnd-rectangular}}\xspace}
\newcommand{\PertNoi}{{\textsf{\PertRnd-noisy}}\xspace}

\newcommand{\MagEucli}{{\textsf{PM-Euclidean distance}}\xspace}
\newcommand{\MagOFd}{{\textsf{PM-obj.func. distance}}\xspace}
\newcommand{\MagSucc}{{\textsf{PM-success rate}}\xspace}
\newcommand{\Mtx}{{\textsf{Mtx}}\xspace}
\newcommand{\MtxIdentity}{{\textsf{\Mtx-identity}}\xspace}
\newcommand{\MtxDiagonal}{{\textsf{\Mtx-random diagonal}}\xspace}

\newcommand{\MtxEuclidean}{{\textsf{\Mtx-Euclidean rotation}}\xspace}

\newcommand{\MtxIncreasingGroupBased}{{\textsf{\Mtx-Increasing group-based}}\xspace}

\newcommand{\AglAdaptive}{{\textsf{$\alpha$-adaptive}}\xspace}
\newcommand{\PhiConstant}{{\textsf{AC-constant}}\xspace}
\newcommand{\PhiRandom}{{\textsf{AC-random}}\xspace}
\newcommand{\PhiTV}{{\textsf{AC-time-varying}}\xspace}
\newcommand{\PhiExtra}{{\textsf{AC-extrapolated}}\xspace}
\newcommand{\MoiBoN}{{\textsf{MoI-best-of-neighborhood}}\xspace}
\newcommand{\MoiFI}{{\textsf{MoI-fully informed}}\xspace}
\newcommand{\MoiRFI}{{\textsf{MoI-ranked fully informed}}\xspace}
\newcommand{\MoiRI}{{\textsf{MoI-random informant}}\xspace}

\newcommand{\TopFC}{{\textsf{Top-fully-connected}}\xspace}
\newcommand{\TopRing}{{\textsf{Top-ring}}\xspace}

\newcommand{\OmegaCons}{{\textsf{IW-constant}}\xspace}
\newcommand{\OmegaLinDec}{{\textsf{IW-linear decreasing}}\xspace}

\newcommand{\OmegaAdapVel}{{\textsf{IW-adaptive based on velocity}}\xspace}

\newcommand{\OmegaRnkBsd}{{\textsf{IW-rank-based}}\xspace}
\newcommand{\OmegaSuccBsd}{{\textsf{IW-success-based}}\xspace}

\newcommand{\ACcomp}{{\textsf{AC}}\xspace}
\newcommand{\PMcomp}{{\textsf{PM}}\xspace}
\newcommand{\TOPcomp}{{\textsf{Top}}\xspace}
\newcommand{\MOIcomp}{{\textsf{MoI}}\xspace}
\newcommand{\POPcomp}{{\textsf{Pop}}\xspace}
\newcommand{\PertRndCS}{{\textsf{perturbation2CS}}\xspace}
\newcommand{\PertInfCS}{{\textsf{perturbation1CS}}\xspace}

\makeatletter
\renewenvironment{cases}[1][\lbrace]{%
	\def\@ldelim{#1}
	\matrix@check\cases\env@cases
}{%
	\endarray\right.%
}
\patchcmd{\env@cases}{\lbrace}{\@ldelim}{}{}
\makeatother

\usepackage[numbers]{natbib}

\usepackage{graphicx}

\usepackage[caption=false,font=normalsize,labelfont=sf,textfont=sf]{subfig}

\usepackage{stfloats}

\begin{document}
\title{Quantifying the Impact of Modules and Their Interactions in the \PSOX Framework}
\author{Christian L. Camacho-Villalón, Ana Nikolikj, Katharina Dost, Eva Tuba, Sašo Džeroski and Tome Eftimov 
    \thanks{This work was supported by the European Union’s Horizon Europe research and innovation program under the Marie Sklodowska-Curie COFUND Postdoctoral Programme grant agreement No.101081355-SMASH and by the Republic of Slovenia and the European Union from the European Regional Development Fund. 
    We also gratefully acknowledge the Slovenian Research and Innovation Agency for funding through program grant P2-0098, project grants J2-4460 and GC-0001, as well as the young researcher grant PR-12897 awarded to Ana Nikolikj. We acknowledge the support of the EC/EuroHPC JU and the Slovenian Ministry of HESI via the project SLAIF (grant number 101254461).}
    \thanks{C.L. Camacho-Villalón and S. Džeroski are with the Department of Knowledge Technologies. A. Nikolikj, T. Eftimov and E. Tuba are with the Computer Systems Department. Both departments are part of the Jo\v{z}ef Stefan Institute in Ljubljana, Slovenia. 
    Email: \texttt{christian.camacho.villalon@ijs.si}.} 
    \thanks{K. Dost is with the School of Mathematics and Statistics at the University of Canterbury in New Zealand.}
    \thanks{Digital Object Identifier XXXXX.YYYYY}
    }

\markboth{Journal of \LaTeX\ Class Files,~Vol.~18, No.~9, September~2020}%
{How to Use the IEEEtran \LaTeX \ Templates}

\maketitle

\begin{abstract}
The \PSOX framework incorporates dozens of modules that have been proposed for solving single-objective \cops using \pso. 
While modular frameworks enable users to automatically generate and configure algorithms tailored to specific optimization problems, the complexity of this process increases with the number of modules in the framework and the degrees of freedom defined for their interaction.
Understanding how modules affect the performance of algorithms for different problems is critical to making the process of finding effective implementations more efficient and identifying promising areas for further investigation.
Despite their practical applications and scientific relevance, there is a lack of empirical studies investigating which modules matter most in modular optimization frameworks and how they interact.
In this paper, we analyze the performance of 1,424 particle swarm optimization algorithms instantiated from the \PSOX framework on the 25 functions in the CEC'05 benchmark suite with 10 and 30 dimensions.
We use \funANV to quantify the impact of modules and their combinations on performance in different problem classes.
In practice, this allows us to identify which modules have greater influence on \PSOX performance depending on problem features such as multimodality, mathematical transformations and varying dimensionality.
We then perform a cluster analysis to identify groups of problem classes that share similar module effect patterns. 
Our results show low variability in the importance of modules in all problem classes, suggesting that particle swarm optimization performance is driven by a few influential modules. 
\end{abstract}


\section{Introduction}
\IEEEPARstart{C}{ontinuous} optimization problems arise in many fields and domains. 
They range from determining parameter values that produce the desired performance of a system (e.g. a simulator) to designing structures that meet safety and performance standards (e.g., a car chassis).
While some \cops can be solved using \textit{exact methods} (for example, the second derivative or Newton's method), many others have complex features that render them unsuitable for exact solutions. Well-known examples of \cops with complex features are those with multimodal, non-differentiable landscapes; a large number of dimensions; and objective functions with no explicit formulation \cite{LueYe1984:Book-linear-nonlinear,AudHar2017:Book-BBO-DFO}. An effective alternative for solving difficult \cops is to use \textit{metaheuristics}, such as \es (\ESs)~\cite{Rec1971PhD,Schwefel1977}, \de (\DE)~\cite{StoPri1997:de}, and \pso (\PSO)~\cite{KenEbe1995pso,EbeKen1995:pso}. Unlike exact methods, metaheuristics are derivative-free optimization techniques that iteratively sample new candidate solutions from the search space to approximate the optimum of the problem.

While metaheuristics require little to no adaptation to work, research has shown that significant performance gains can be achieved by carefully selecting the algorithm components used in the implementation and fine-tuning their parameter values~\cite{BirStuPaqVar02:gecco,Birattari09tuning,LopStu2010:gecco}. 
Consequently, a great deal of research has focused on improving performance through manual adjustments to various metaheuristics, while much less has been devoted to systematically investigating why some algorithms produce good results for certain problem classes but not others~\cite{Hoo1996joh,GarGutMol2017}.
Recent work addresses this gap by dedicating more effort to provide useful explanations of algorithm performance via theoretical and experimental studies (see, e.g., \cite{Nikolikj2025:benchmarking}).
Moreover, the developments on automatic algorithm configuration and the widespread use of machine learning (ML) to analyze algorithms' performance data have made increasingly efficient to adapt metaheuristic implementations to specific problems, as well as performing fair and reproducible comparisons of different algorithmic variants to assess their strengths and weaknesses in specific scenarios~\cite{DoeWanYe2018:IOHprofiler,BarDoeBer2020:BenchmarkingInOptimization}.
In recent years, researchers have also begun to closely examine the use of automatic methods to develop high-performance implementations from reusable algorithm components, as well as working on the challenges of creating fully automated algorithm pipelines~\cite{CamStuDor2023:IC:disNewMH}.

Despite the advances in the field and the numerous tools available nowadays to implement and study metaheuristics, there is still a mismatch between the extremely large number of algorithm variants available in the literature and the relatively small number of experimental and theoretical studies providing clear guidelines on which algorithms work best for which problems and under which conditions.
Selecting an optimization algorithm can be quite challenging, particularly for inexperienced users and those facing problems that differ greatly from those they have encountered before.
Moreover, if the selected approach does not produce the desired results, the user is faced with an overwhelming number of design alternatives to improve implementation performance. 

\textbf{Our contribution}: This paper sheds light on the importance and interaction of fundamental algorithm components in \PSO, with the aim of assisting users in selecting and adapting a \PSO algorithmic variant.
To do so, we examine the \PSOX framework~\cite{CamDorStu2022:tec}, a modular implementation of \pso composed of 
11 modules and 58 implementation options, spanning over 25 years of research on \PSO. After drawing on performance data from 1,424 \PSO variants generated with \PSOX, we applied a functional analysis of variance (\fANV)~\cite{HutHooLey2014icml,RijHut18:sigkdd} to quantify the influence of individual components and their interactions on the performance of the 25 functions belonging to the CEC'05 "Special Session on Single Objective Real-Parameter Optimization"~\cite{SugHanLia2005cec}. 
This paper adds to the previous data-driven studies examining modules importance on modular optimization frameworks, thus filling an important gap in the literature, since \PSOX has been so far excluded from such studies.

Unlike traditional assessments of module performance, which focus on the impact of a single module or a limited number of module combinations and neglect broader interactions, in this study we consider pairwise and triple interactions. 
Our results shows that \PSO performance heavily depends on the presence and interaction of \OmgCS, \rndMtx and \DNPP modules, although the influence of the latter is weaker. 
Other modules, such as the \AcCof, \Topo, \Moi---despite being the object of much research~\cite{BanVinAnY2007:natc-part1,BanVinAnY2008:natc-part2,BonMic2017:mit} and even controversy~\cite{Eng2013:brics}---contribute only marginally to \PSOX algorithms performance.

We also study how different problem classes cluster based on module effects, in order to establish relationships between the features of high-level problem classes, such as multimodality, separability, and mathematical transformations, and the usage of specific modules in implementation.
We found that the module that most strongly influences \PSOX performance on most functions is \rndMtx, while \OmgCS dominates in a few number of cases.
The effect of single modules is strongest on the 10-dimensional problems and attenuates slightly on the 30-dimensional problems. 
As the number of dimensions increases, the single influence effect of \rndMtx and \OmgCS diminishes and pairwise interactions involving the \DNPP module, which allows to balance out these two modules, become more relevant.

The rest of the paper is structured as follow.
\Sect\ref{sec:Background and Related Work} reviews background and related work; \Sect\ref{sec:methodology} presents our module-importance methodology; \Sect\ref{sec:experimental_design} details the experimental setup; \Sect\ref{sec:results} reports results; and \Sect\ref{sec:Conclusions} concludes with limitations and future work.

\section{Background and Related Work}\label{sec:Background and Related Work}
In this work, we consider single-objective \cops (COPs).
In a single–objective COP, the goal is to find a real-valued vector 
$ \vec{x}^\ast \in \mathbb{R}^D $
that minimizes an objective function 
$ f : \mathbb{R}^D \rightarrow \mathbb{R}, $
such that $\vec{x}^\ast = \arg \min_{\vec{x} \in \mathbb{R}^D} f(\vec{x})$, 
where $D$ denotes the dimensionality of the decision space. 
We also consider the case where $f(\cdot)$ is non-linear, multimodal, and possibly non-separable, making it unfeasible to analytically compute gradients or closed-form solutions.
To address this type of problems, metaheuristics are the method of choice. 

\subsection{Automatic Algorithm Design}
\label{sec:Automatic algorithm design}
New variants of metahueristic algorithms are usually proposed in an incremental manner, i.e., one or just a few at a time, and their design is the result of a manual process guided by the intuition and expertise of the algorithm designers. This approach to design algorithms, while successful in small number of cases, is often subjective, time-consuming, and error-prone.
To address these issues, the automatic design approach leverages 
the development of component-based optimization frameworks
and 
the advances in automatic algorithm configuration%
~\cite{StuLop2019hb}.
Component-based optimization frameworks, such as \modCMAES~\cite{deNVerWan2021:gecco-CMAESframework}, \modDE \cite{VerCarKon2023:autoDE:gecco}, \PSOX~\cite{CamDorStu2022:tec} and \metafor~\cite{CamDorStu:25:telo}, provide users with a flexible way to generate many different algorithmic implementations from a discreet  sets of algorithm components (modules), which are interchangeable in the algorithm design and responsible for specific behaviors.
On the other hand, automatic algorithm configuration tools (AACTs), e.g., \irace~\cite{LopDubPerStuBir2016irace}, \paramils~\cite{HutHooLeyStu2009jair}, and \smac~\cite{HutHooLey2011lion}, perform the task of trying different combinations of algorithm components and assessing their performance on different problems.


\subsection{The \PSOX Framework}
\label{sec:the psox framework}
The \PSOX framework~\cite{CamDorStu2022:tec} is a component-based (i.e., modular) implementation of the \pso algorithm~\cite{KenEbe1995pso,EbeKen1995:pso}. The framework incorporates a wide range of implementations options inspired by various state-of-the-art \PSO variants, where key design choices have been translated into distinct modules.
The \PSOX architecture decomposes the algorithm into interchangeable modules that govern, for example, initialization, velocity update rule, position update rule, topology and model of influence control strategies, etc. 
At its core, \PSOX employs a generalized velocity update rule that unifies and extends multiple \PSO variants, allowing the inclusion and exclusion of different components, such as inertia and acceleration coefficients, perturbation strategies, random matrices, angle-based rotations, among others. 
This formulation enables flexible algorithm design of both classical and new \PSO dynamics under a common algorithmic template. 
Moreover, by pairing it with an \AACT, \PSOX can be used to systematically explore design alternatives and assess the impact that individual components and their interactions have on the algorithms across different optimization problems.
A summary of the main algorithm components implemented in the \PSOX framework is given in \Tbl~\ref{tab:psox_components}. For further details about \PSOX, we refer the reader to \cite{CamDorStu2021:posx-supp,CamDorStu2022:tec}.
In the remainder, we use a \textsf{sans-serif} font to indicate both the modules implemented in the \PSOX framework and their available options.

\begin{table*}[t]
\centering
\footnotesize 
\caption{
Main algorithmic components implemented in the \PSOX framework. Each component defines a modular dimension in the algorithm template and can be combined with others to instantiate a complete \PSO variant.}
\label{tab:psox_components}
\begin{tabularx}{\textwidth}{@{}p{2.8cm} X@{}}
\toprule
\textbf{Algorithm component} & \textbf{Description} \\
\midrule

Generalized Velocity\newline Update Rule (GVUR)
&
Core kinematic rule unifying \PSO formulations.  
Particles' positions ($\vec{x}^{i}$), which represent solutions to the optimization problem, are updated as $\vec{x}^{i}_{t+1}=\vec{x}^{i}_{t}+\vec{v}^{i}_{t+1}$.
The computation of the velocity vector ($\vec{v}^{i}_{t+1}$) is done using using a GVUR defined as: $\vec{v}^{\,i}_{t+1} = \omega_1\,\vec{v}^{\,i}_t + \omega_2\,\DNPP(i,t) + \omega_3\,\PertRnd(i,t),$, where  $\boldsymbol{v}^{i}_{t}$ is the inertial term, \DNPP and \PertRnd are two modular components, and $\omega_1$, $\omega_2$, and $\omega_3$ are three real-value parameters.\\
\cmidrule{1-2}
\DNPP 
& The distribution of Next Possible Positions (\DNPP) defines the geometric distribution used to compute particle displacements, influencing rotational invariance and exploration amplitude. For example, \DNPPRect uses a hyper-cubic distribution, the \DNPPSphe uses hyper-spherical distribution and \OperatorN, a multivariate Gaussian distribution.\\
\cmidrule{1-2}
\PertRnd (\PertRndCS)
& Applies optional non-informative stochastic perturbations independently of swarm and personal information to increase exploration. Two main implementation options are \PertRect and \PertNoi, both sampling from random uniform distribution.\\
\cmidrule{1-2}

\PertInf (\PertInfCS)
& Applies optional informative perturbations centered on a direction of interest within the \DNPP component to enhance exploitation focused on local refinement around know solutions. Some implementation options include, \PertGau, \PertLev and \PertUni.\\
\cmidrule{1-2}

\PMcomp
& Defines the perturbation magnitude (\PMcomp) used by \PertRnd and \PertInf.  
Main options include \MagEucli and \MagOFd, which adjust the perturbation based on spatial or fitness improvement, and \MagSucc, which increases/decreases the magnitude according to the rate of successful perturbations.  These strategies are only active when \PertInf or \PertRnd modules are enabled.\\
\cmidrule{1-2}

\OmgCS
& Controls particles' inertia ($\omega_1$) in the GVUR. Its value can be
computed using a variety of strategies, ranging from time-varying (e.g., \OmegaLinDec) to adaptive updates (e.g., \OmegaAdapVel).\\
\cmidrule{1-2}

\AcCof (\ACcomp)
& Controls for the cognitive and social acceleration coefficients $(\varphi_1,\varphi_2)$.  
It provides a balance between self-reinforcement and cooperation. Main options include \PhiExtra and \PhiTV, which adapt the coefficients dynamically to iteration and quality differences. \\
\cmidrule{1-2}

\rndMtx (\Mtx)
& This component is only available for the \DNPPRect or \DNPPSphe modules. It generates random transformation matrices used to rotate or scale displacement vectors, such as \MtxDiagonal and \MtxEuclidean (with angle $\alpha$ following constant, Gaussian, or adaptive schedules), and progressively diagonalized group-based matrices, such as \MtxIncreasingGroupBased.\\
\cmidrule{1-2}

\Topo (\TOPcomp)
& Defines the neighborhood structure $\mathcal{N}_i$ that governs information exchange.  
Different topologies balance exploration and exploitation through connectivity, being \TopRing the less connected and \TopFC the more connected.\\
\cmidrule{1-2}

\Moi (\MOIcomp)
& Specifies which solutions are chosen from $\mathcal{N}_i$ and how their influence on a particle's movement will be weighted, e.g., \MoiBoN, \MoiFI (averaged), \MoiRFI, or \MoiRI.\\
\cmidrule{1-2}

\Pop (\POPcomp)
& Controls swarm size dynamics and initialization.  
For example, \PopTV and \PopIncre strategies add or remove particles depending on search progress. Two intialization schemes are considered: \InitRandom, which uses random sampling, and \InitHorizontal, which combines random sampling with horizontal learning toward the global best.\\

\bottomrule
\end{tabularx}
\end{table*}

\subsection{Functional analysis of variance}
\label{sec:f-ANOVA}
Functional analysis of variance (\fANV)~\cite{HutHooLey2014icml,RijHut18:sigkdd} is a variance-decomposition technique originally used in hyperparameter optimization for machine learning.
\fANV is used to explain how much each factor, and combinations of factors, contributes to variability in an outcome across a set of experiments. 
Given observed performance values over many configurations, \fANV marginalizes over all other factors to estimate the effect of a single factor (main effect) and of higher-order interactions (pairs, triples, etc.). 
The resulting effect contributions are additive, up to estimation error, and sum to the total observed variance, yielding interpretable importance scores that rank factors and interactions by explanatory power.
To apply the \fANV technique, one must provide a design table that includes the factors and their respective levels for each run, as well as the corresponding responses. \fANV then determines the relative contributions, which allows to identify dominant drivers, as well as synergistic or antagonistic interactions and context dependencies (e.g., by dataset or task).

\subsection{Related Work}
\label{sec:related-work}
Previous research on the performance of modular optimization frameworks has mainly studied module importance via analyses of top-performing configurations, often relying on frequency counts of selected components. For example, in studies on \modCMAES~\cite{deNVerWan2021:gecco-CMAESframework} and \modDE \cite{VerCarKon2023:autoDE:gecco}, the automatic configuration tool \irace was used to identify “elite” configurations, and the importance of a module was assessed by its frequency among these elites. 
To capture interaction effects, modules were incrementally added to the configuration space and the change in elite frequencies was observed. 
This type of studies provide a coarse view of module contributions and do not systematically contrast module importance across different problem classes. 

Several complementary techniques for assessing hyperparameter and module importance have been proposed in the broader optimization and machine learning communities. 
These include forward-selection methods~\cite{HutHooLey2013lion}, performance-influence models~\cite{Siegmund2015}, ablation analysis~\cite{BieLinEggFraFawHoo2017}, and \funANV~\cite{HutHooLey2014icml,RijHut18:sigkdd}, each offering a different perspective.
Ablation, for instance, disables or replaces one component at a time to measure the performance drop, whereas \fANV partitions performance variance across algorithms components and their combinations. 

Researchers have also explored learning models to predict effective module choices from problem features. In \cite{Prager2020}, \citeauthor{Prager2020} use a classifier-chain approach to predict each module’s optimal setting based on landscape characteristics, implicitly modeling dependencies between modules but without quantifying which interactions are most important. Similarly, in \cite{KosVerDze22:gecco,Kostovksa2025}, \citeauthor{KosVerDze22:gecco} train separate machine learning models to recommend the best component for each \modCMAES module given problem feature values. 
While these approaches acknowledge that problem properties influence module efficacy, they generally treat modules independently and thus overlook explicit module-to-module interaction effects. In a recent study \cite{SteVerKon24:xai-heu},  \citeauthor{SteVerKon24:xai-heu} applied explainable AI techniques to modular algorithms, but focused only on individual module importance and ignored higher-order interactions.

The only work to date that explicitly quantifies variance contributions of both individual modules and their combinations in modular optimization frameworks is the one reported in \cite{NikKosVer24:cec,Nikolikj2025:sec}. 
In \cite{NikKosVer24:cec}, \citeauthor{NikKosVer24:cec} applied \fANV to \modCMAES and \modDE variants to measure main and pairwise/triple interaction effects, and in \cite{Nikolikj2025:sec}, they studied the alignment of clusters of problem classes with similar module interaction patterns with that of clusters based on high-level problem characteristics.
So far, no comparable variance-decomposition analysis has been performed for the \PSOX framework under different problem landscapes. Our work fills this gap by providing the first problem-specific module importance and interaction analysis for PSO. 

\section{Methodology for Quantifying Module Effects}
\label{sec:methodology}
To analyze the role of \PSOX modules across different problem landscapes, we employ a two-step methodology grounded in \fANV. 
First, for each problem class (see \Sect\ref{sec:Benchmark Set of Functions}), we construct a dataset of all \PSOX variants evaluated on that function. Each variant is encoded by its module choices as categorical features and its performance on the function as the target variable. We then apply \fANV to each dataset to quantify how much variance in performance is explained by each module alone (individual effects), by each pair of modules (pairwise interaction effects), and by each triple of modules (triplets interaction effects). This produces, for every problem class, a vector of effect contributions of dimension $n + \binom{n}{2} + \binom{n}{3} = 92$ (with $n=8$ modules in \PSOX).
Note that although \PSOX has 11 modules in total, to keep the number of algorithm variants manageable,
we have fixed the following parameters: $\omega_2=1$ and $\omega_3=1$, in the GVUR, $\PMcomp = 0.5$, in both \PertRnd and \PertInf, and \PopConst with \texttt{pop}$_{size} = 20$, in \POPcomp.

\subsection{\fANV for Module Interaction Analysis}
\label{sec:f-ANOVA for module interaction analysis}
Given that \PSOX can instantiate a large number of algorithmic configurations through different combinations of modules (e.g., velocity update, topology, inertia control, perturbation strategy), \fANV enables us to disentangle the influence of each module on the observed algorithm performance.
Let  $g(\vec{m})$ denote the expected performance of a \PSOX configuration defined by the module vector 
$\vec{m} = (m_1, \dots, m_q)$, 
where each $m_i$ represents a categorical or numerical hyperparameter controlling a specific module or implementation option.
Under the assumption that $g(\vec{m})$ is square-integrable, it can be decomposed into a sum of main and interaction effects as follows:
\begin{multline}
g(\vec{m}) = g_0 
  + \sum_{i=1}^{q} g_i(m_i)
  + \sum_{i<j} g_{i,j}(m_i, m_j)
  + \dots\\
  + g_{1,\dots,q}(m_1, \dots, m_q),
\end{multline}
where $g_0$ is the mean performance across all configurations, 
$g_i(m_i)$ is the main effect of module $i$, and higher-order terms 
$g_{i,j}(\cdot), g_{i,j,k}(\cdot)$, $g_{i,\dots,q}(\cdot)$, denote the interaction effects between modules or parameter groups.  
Analogously, the total variance of $g(\vec{m})$ can be written as:
\begin{equation}
\mathbb{V}[g(\vec{m})] = 
  \sum_{U \subseteq \{1, \dots, q\}} \mathbb{V}_U,
\end{equation}
where $ \mathbb{V}_U = \mathrm{Var}\!\left[ g_U(\vec{m}_U) \right] $
represents the variance contribution of subset $U$ (i.e., a group of modules).  
Then, the relative importance of $U$ is given by:
\begin{equation}
\mathbb{I}_U = 
  \frac{\mathbb{V}_U}{\mathbb{V}[g(\vec{m})]},
\end{equation}
with $\sum_U \mathbb{I}_U = 1$.  
High values of $\mathbb{I}_U$ indicate that the corresponding module or module combination has a strong effect on the optimization performance.

Computing these variance components directly is infeasible, as \PSOX embodies a high-dimensional combinatorial design space that mixes large number of both categorical module selections and continuous control parameters.  
Instead, a surrogate regression model 
is trained on empirical results obtained from multiple \PSOX configurations. Following the approach of~\citeauthor{HutHooLey2014icml}~\cite{HutHooLey2014icml}, we employ a random-forest regressor to approximate the performance function and to efficiently estimate the marginal contributions of each module and their interactions.
The estimated variance components and their normalized importance allow quantifying how much each \PSOX module (and its interaction with others) contributes to overall performance variability. These results can be aggregated per module category (e.g., topology, inertia, perturbation mechanism) and compared across problem classes to systematically identify dependencies between algorithmic design choices and problem landscape features.

\subsection{Problem Clustering Based on Module Importance}
We use the variance effect vectors to identify groups of problem classes with similar module importance patterns. We treat each problem class in the \fANV result as an $92$-dimensional embedding (i.e., vector of meta-features) of that class. Clustering is then performed on these 25 embeddings, one per CEC’05 function, to discover clusters of problems that "activate" \PSOX modules in similar ways. 
Because we want to preserve all information from the variance contributions, clustering is conducted directly in the 92-dimensional space without dimensionality reduction.
We explore various cluster counts and linkage criteria, using the Silhouette coefficient to guide the selection of an appropriate number of clusters. 
After determining the clustering, we examine the composition of each cluster in terms of known problem characteristics. In particular, we compare our data-driven grouping with the established categories of the CEC’05 benchmark. This comparison reveals the extent to which common problem features (such as modality or separability) coincide with similar module importance profiles.

\section{Experimental Design}
\label{sec:experimental_design}
To collect \PSOX performance data, we used the CEC'05 benchmark suite~\cite{SugHanLia2005cec}. 
The CEC'05 suite remains a useful benchmark to test metaheuristic algorithms due to the complexity and diversity of its problems. Below, we provide further details on the CEC'05 suite, the \PSOX configuration space, and the datasets used by \fANV.

\subsection{Benchmark Set of Functions}
\label{sec:Benchmark Set of Functions}
The CEC'05 benchmark suite~\cite{SugHanLia2005cec} comprises 25 problem that can be separated into three main classes based on their high-level features:
unimodal functions ($f_1$-$f_5$), basic multimodal functions ($f_6$-$f_{12}$), expanded multimodal functions ($f_{13}$-$f_{14}$), and hybrid composition functions ($f_{15}$-$f_{25}$).
In the CEC'05 benchmark, across all problem classes, there are a number of properties that make the suite highly heterogeneous. For example, there are: \textit{non-separable functions}, that is, functions that cannot be optimized dimension by dimension; \textit{mathematical transformations}, namely functions with shifted, rotated and scaled search space; and \textit{inaccessible global optima} that are located in narrow basins, flat regions, the bounds of the search space, and outside the initialization range.
Table~\ref{tbl:CEC05} depicts the 25 functions, their properties, domains, and global optima.
We consider the 10 and 30 dimensional versions of the problems, i.e.,  $D \in \{10,30\}$.

\begin{table*}[t]
\centering
\footnotesize
\caption{CEC’05 Benchmark Problems: Properties, Domains, and Global Optima}
\label{tbl:CEC05}
\begin{tabular}{clllcc}
\toprule
\textbf{Func} & \textbf{Name} & \textbf{Type} & \textbf{Properties} & \textbf{Domain} & \textbf{Optimum} \\
\midrule
$f_{1}$  & Shifted Sphere                        & Unimodal & Separable, shifted               & $[-100,100]^D$   & 0 \\
$f_{2}$  & Shifted Schwefel’s 1.2                & Unimodal & Non-separable, shifted            & $[-100,100]^D$   & 0 \\
$f_{3}$  & Shifted Rotated High Cond. Elliptic   & Unimodal & Non-separable, rotated, shifted   & $[-100,100]^D$   & 0 \\
$f_{4}$  & Shifted Schwefel’s 1.2 + Noise        & Unimodal & Non-separable, noisy              & $[-100,100]^D$   & 0 \\
$f_{5}$  & Schwefel’s 2.6 (bounds)               & Unimodal & Non-separable, optimum on bounds  & $[-100,100]^D$   & 0 \\
\midrule
$f_{6}$  & Shifted Rosenbrock                    & Multimodal & Non-separable, narrow valley    & $[-100,100]^D$   & 0 \\
$f_{7}$  & Shifted Rotated Griewank (no bounds)  & Multimodal & Non-separable, rotated          & $[-600,600]^D$   & 0 \\
$f_{8}$  & Shifted Rotated Ackley (bounds)       & Multimodal & Non-separable, rotated, bounds  & $[-32,32]^D$     & 0 \\
$f_{9}$  & Shifted Rastrigin                     & Multimodal & Separable, many local optima    & $[-5,5]^D$       & 0 \\
$f_{10}$ & Shifted Rotated Rastrigin             & Multimodal & Non-separable, rotated, many optima & $[-5,5]^D$       & 0 \\
$f_{11}$ & Shifted Rotated Weierstrass           & Multimodal & Non-separable, fractal landscape& $[-0.5,0.5]^D$   & 0 \\
$f_{12}$ & Schwefel’s 2.13                       & Multimodal & Non-separable, shifted          & $[-\pi,\pi]^D$   & 0 \\
\midrule
$f_{13}$ & Expanded Griewank+Rosenbrock ($f_{8}$,$f_{2}$)   & Multimodal & Non-separable, expanded hybrid  & $[-3,1]^D$       & 0 \\
$f_{14}$ & Shifted Rotated Scaffer’s $f_{6}$          & Multimodal & Non-separable, rotated          & $[-100,100]^D$   & 0 \\
\midrule
$f_{15}$ & Hybrid Composition 1                  & Hybrid    & Mixed, partly separable          & $[-5,5]^D$       & 0 \\
$f_{16}$ & Rotated Hybrid 1                      & Hybrid    & Non-separable, rotated           & $[-5,5]^D$       & 0 \\
$f_{17}$ & Rotated Hybrid 1 + Noise              & Hybrid    & Non-separable, noisy             & $[-5,5]^D$       & 0 \\
$f_{18}$ & Rotated Hybrid 2                      & Hybrid    & Non-separable, traps, flat regions& $[-5,5]^D$      & 0 \\
$f_{19}$ & Rotated Hybrid 2 (narrow basin)       & Hybrid    & Non-separable, narrow basin      & $[-5,5]^D$       & 100 \\
$f_{20}$ & Rotated Hybrid 2 (opt on bounds)      & Hybrid    & Non-separable, optimum on bounds & $[-5,5]^D$       & 0 \\
$f_{21}$ & Rotated Hybrid 3                      & Hybrid    & Non-separable, rotated           & $[-5,5]^D$       & 200 \\
$f_{22}$ & Rotated Hybrid 3 (ill-conditioned)    & Hybrid    & Non-separable, ill-conditioned   & $[-5,5]^D$       & 300 \\
$f_{23}$ & Non-continuous Rotated Hybrid         & Hybrid    & Non-separable, non-continuous    & $[-5,5]^D$       & 300 \\
$f_{24}$ & Rotated Hybrid 4                      & Hybrid    & Non-separable, rotated, complex  & $[-5,5]^D$       & 200 \\
$f_{25}$ & Rotated Hybrid 4 (no bounds)          & Hybrid    & Non-separable, opt outside init  & $[-5,5]^D$       & 200 \\
\bottomrule
\end{tabular}
\end{table*}

\subsection{Configuration Space}
To systematically generate the \PSO variants considered in our study, we took the Cartesian product of eight modules with a total of 26 implementation options (see Table~\ref{tbl:PSO-X_modules}), resulting in a total of 1,424 variants. Each algorithmic variant was executed independently 10 times on each of the 25 problem classes from the CEC'05 benchmark suite. The evaluation budget was set to $5000D$ function evaluations, where $D$ is the number of dimension of the optimization problem. Performance was assessed based on the distance between the best-found solution and the global optimum (\textit{distance}). To ensure numerical stability and meaningful comparisons, distances smaller than $10^{-9}$ were capped at this threshold. For each problem instance, the median distance across the 10 runs was taken as the final performance measure. These values were then log-transformed using base 10. As a result, the~\textit{distance} metric has a lower bound of $10^{-9}$, with lower values indicating better optimization performance.

\begin{table*}[t]
\centering
\caption{The 8 \PSOX modules and 26 implementation options considered in this work. The implementation options are number coded as they are in the \PSOX framework. The default values for the parameters associated with specific implementation options are shown in parentheses.
}
\label{tbl:PSO-X_modules}
\footnotesize  
\begin{tabular}{>{\raggedright\arraybackslash}p{2.5cm} >{\raggedright\arraybackslash}p{14cm}}
    \toprule
    \textbf{Module} & \textbf{Implementations options and parameter vaslues} \\
    \midrule
    \DNPP & 0=\DNPPRect, 1=\DNPPSphe, 2=\DNPPAddStoch (\textsf{parm\_r} = 0.5) \\\cmidrule{1-2}
    \AcCof & 0=\PhiConstant ($\phi_1= 1.4$, $\phi_2 = 1.4$), 1=\PhiRandom ($\phi_{1,t_0} = 2.4$, $\phi_{1,t_{max}} = 0.5$, $\phi_{2,t_0} = 0.5$, $\phi_{2,t_{max}} = 2.4$) \\\cmidrule{1-2}
    \Topo & 0=\TopRing, 1=\TopFC \\\cmidrule{1-2}
    \Moi & 0=\MoiBoN, 1=\MoiFI \\\cmidrule{1-2}
    \rndMtx & 0=\MtxIdentity, 1=\MtxDiagonal, 4=\MtxEuclidean (\AglAdaptive, \textsf{par\_alpha} = 30, \textsf{par\_beta} = 0.01), 6=\MtxIncreasingGroupBased and \none \\\cmidrule{1-2}
    \OmgCS & 0=\OmegaCons ($\omega_1 = 0.0$), 0=\OmegaCons ($\omega_1 = 0.75$), 12=\OmegaAdapVel ($\lambda 0.5$, $\omega_{t_0} = 0.15$, $\omega_{t_{max}} = 0.95$), 14=\OmegaRnkBsd ($\omega_{t_0} = 0.15$, $\omega_{t_{max}} = 0.95$), 15=\OmegaSuccBsd ($\omega_{t_0} = 0.15$, $\omega_{t_{max}} = 0.95$)\\\cmidrule{1-2}
    \PertInfCS & 0=\none, 1=\PertGau (\MagSucc, $\PMcomp = 0.5$, $\textsf{success} = 40$, $\textsf{failure} = 20$) \\\cmidrule{1-2}
    \PertRndCS & 0=\none, 1=\PertRect (\MagSucc, $\PMcomp = 0.5$, $\textsf{success} = 40$, $\textsf{failure} = 20$) \\
    \bottomrule
\end{tabular}
\vspace{1mm}
\end{table*}

\subsection{\fANV Datasets and Hierarchical Clustering}
Since this study aims to quantify the importance of individual modules for each of the 25 problem classes, we organize the data into 25 separate datasets, one for each problem class. Each dataset contains 1,424 algorithm variants (data instances), where each instance is described by eight module settings (features), and the target variable corresponds to the performance of the variants on the respective problem class. 
Running \fANV on each dataset individually results in 25 vector representations, each capturing the module effects through $8 + \binom{8}{2} + \binom{8}{3}$ terms. To identify similarities among problem classes based on these effects, we apply Hierarchical Clustering (HC)~\cite{Mul11:hc}. Given the relatively small dataset (25 instances) and our focus on interpretability, HC was chosen for its ability to provide a clear and transparent clustering process via a dendrogram, which visually illustrates how clusters are formed and merged at each stage.
To determine the most suitable number of clusters, we conduct a grid search over the clustering algorithm’s hyperparameters and evaluate the results using the Silhouette coefficient~\cite{Rou87:silhouettes}. This metric, ranging from -1 to 1, reflects the quality of clustering—where higher values indicate more compact and well-separated clusters. 


\section{Results and Discussion}
\label{sec:results}

We organize our results and discussion into four parts. First, we examine the overall performance distribution of \PSOX algorithm variants on each problem class. 
Second, we analyze the cumulative variance contribution of ranked modules effects, that is, $8$ individual modules effects, plus $\binom{8}{2} = 28$ pairwise modules effects, plus $\binom{8}{3} = 56$ triple modules effects.
Third, we present a clustering of problem classes based on module effect vectors and examine how these clusters correspond to the features of CEC'05 problems. 
To understand how certain module options and their interactions affect the performance of \PSOX in different clusters, we use marginal performance profiles (i.e., lower distance to the optimum) and discuss representative cases.
Due to space limitations, only representative plots are presented here. However, the full set of plots, raw and processed data, and scripts used for processing and visualization are available as supplementary material for this article. 

\subsection{Algorithm Performance Distribution}
\label{sec:Algorithm performance distribution}

\begin{figure*}[t]
    \centering
    \subfloat[10$D$ problems.]{
        \includegraphics[width=0.8\textwidth]{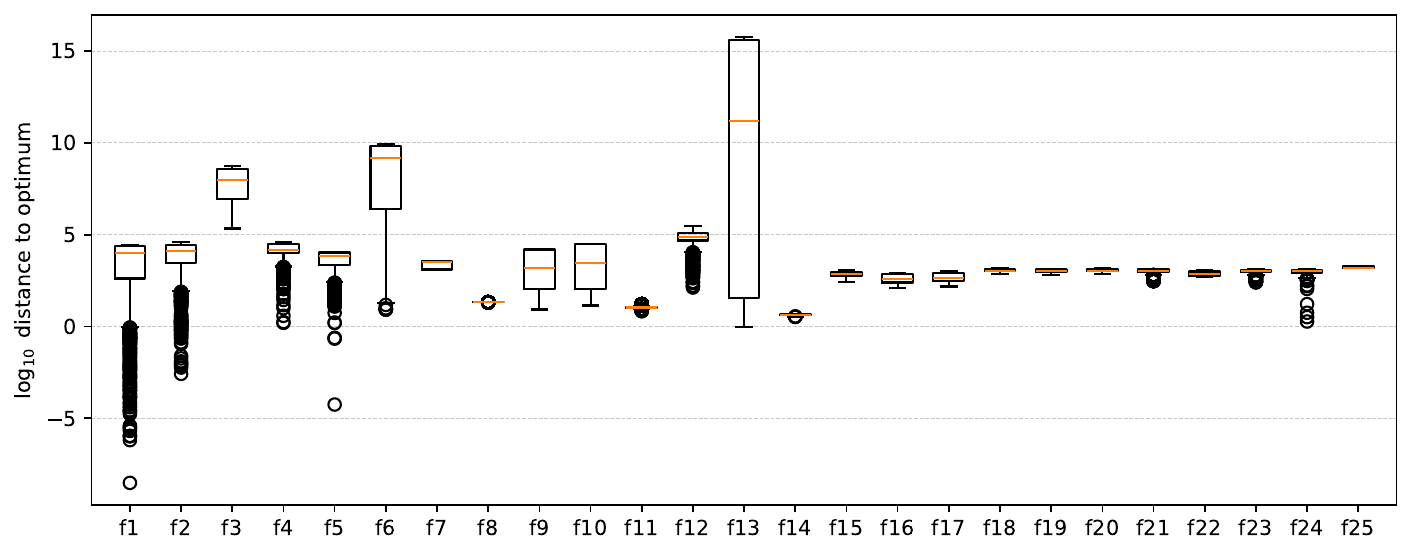}\label{fig:performance10}}
    \quad
    \subfloat[30$D$ problems.]{
        \includegraphics[width=0.8\textwidth]{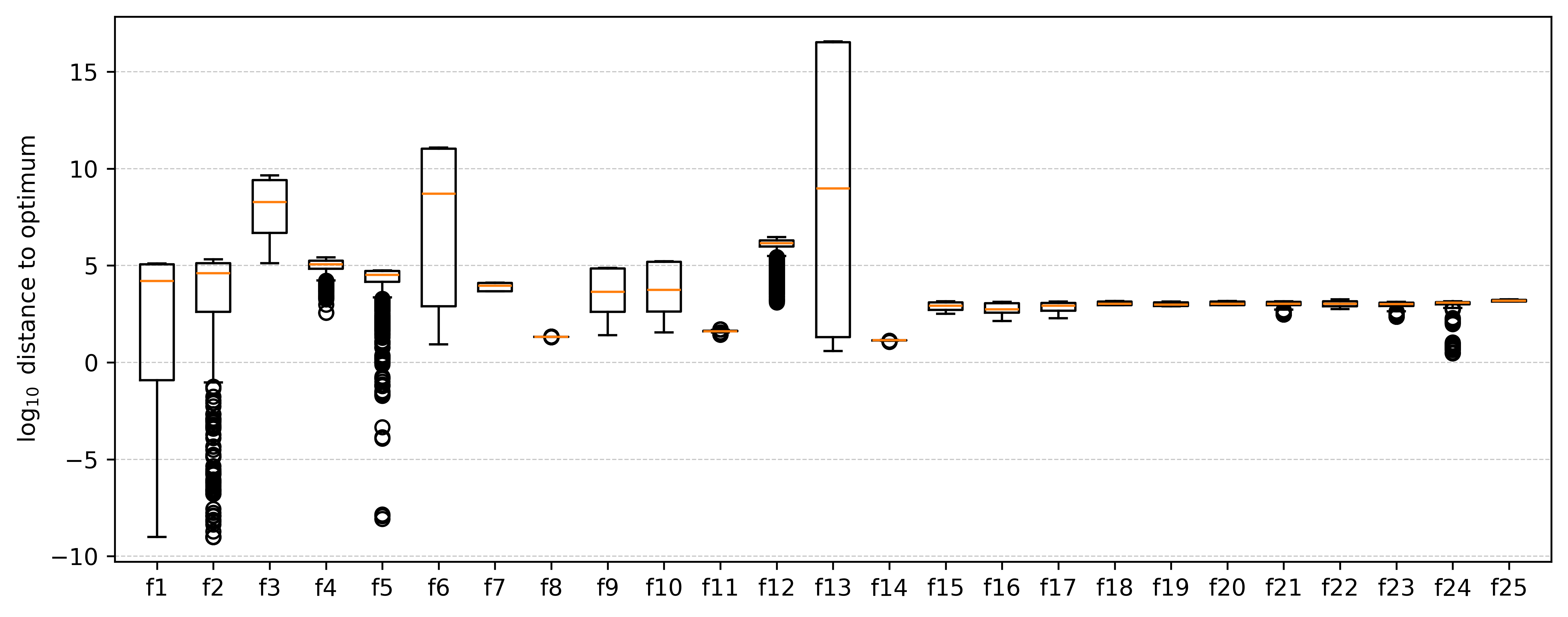}\label{fig:performance30}}
    \caption{Performance distribution of 1,424 \PSOX variants on 25 problem classes comprising the CEC'05 suite. The performance values (i.e., the absolute difference between the best-found solution (within $5,000D$ evaluations) and the global optimum) are presented on a logarithmic scale.}
\end{figure*}

We begin by evaluating the performance distribution of the $1,424$ \PSOX variants on each of the 25 benchmark functions. Performance is measured using the log-transformed \textit{distance} metric, which is the difference between the best objective value found by a variant (within the budget of $5000D$ evaluations) and the global optimum. Lower \textit{distance} values indicate better outcomes, with a minimum possible value of $-9$ (i.e., reaching $10^{-9}$ accuracy). The performance distribution of all \PSOX variants is show as boxplots for each of the 10$D$ (Figure~\ref{fig:performance10}) and 30$D$ (Figure~\ref{fig:performance30}) problems---functions $f_1$ through $f_{25}$. 
The spread of each boxplot reflects the variability in the performance achieved by \PSOX algorithms using different modules.
Classes with substantial boxplot spread (i.e., $f_1$, $f_2$, $f_3$, $f_5$, $f_6$, $f_9$, and $f_{13}$) show a large gap between the best and worst \PSOX variants, indicating high sensitivity to the choice of modules and signaling significant potential gains by adapting the algorithm through redesign and parameter configuration. 
In contrast, several problem classes (notably $f_7$, $f_8$, $f_{11}$, and most functions from $f_{15}$–$f_{25}$) exhibit very tight performance distributions where nearly all variants perform similarly. In these cases, even substantial changes in algorithm design yield little difference in outcome, implying that the problem is either easy enough that most configurations succeed or intrinsically difficult such that all configurations struggle more or less equally.

The influence of problem dimensionality on performance variability does not seem to be particularly large when comparing results at $D=30$. While there is some increase in the variability of the results, most PSO-X configurations are capable to scale to higher-dimensional versions of the problems. The decrease in solution quality is more noticeable for functions $f_4$-$f_{10}$, $f_{12}$ and $f_{13}$, where the gap between the best and worst variants widens. However, for functions $f_1$, $f_2$ and $f_5$, there is a higher number of \PSOX configurations reaching better, near-optimal solutions, and for function $f_{15}$-$f_{25}$, the performance remains similar. This trend mirrors observations in other modular algorithm studies (e.g., \cite{NikKosVer24:cec,Nikolikj2025:sec}), where performance dispersion grows with problem complexity. 
While most hybrid problems in the CEC’05 suite are relatively configuration-insensitive (most \PSO-X variants perform similarly), unimodal and multimodal problems offer substantial room for improvement through appropriate module selection---especially as dimensionality increases.


\subsection{Cumulative Module Importance}
\label{sec:Cumulative module importance}

\begin{figure*}[t]
    \centering
    \subfloat[10$D$ problems.]{
        \includegraphics[width=0.45\textwidth]{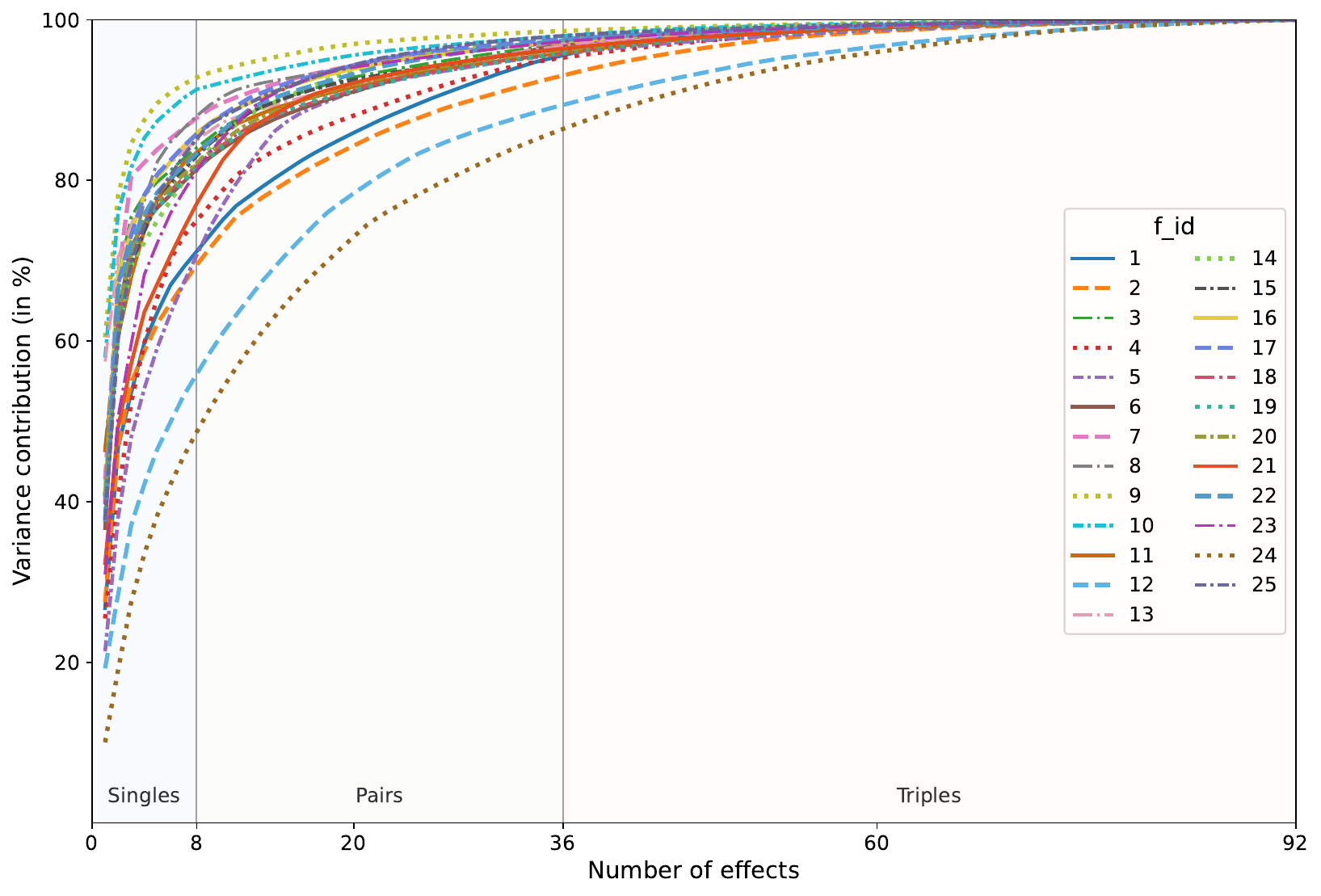}\label{fig:cumvar10}}
    \quad
    \subfloat[30$D$ problems.]{
        \includegraphics[width=0.45\textwidth]{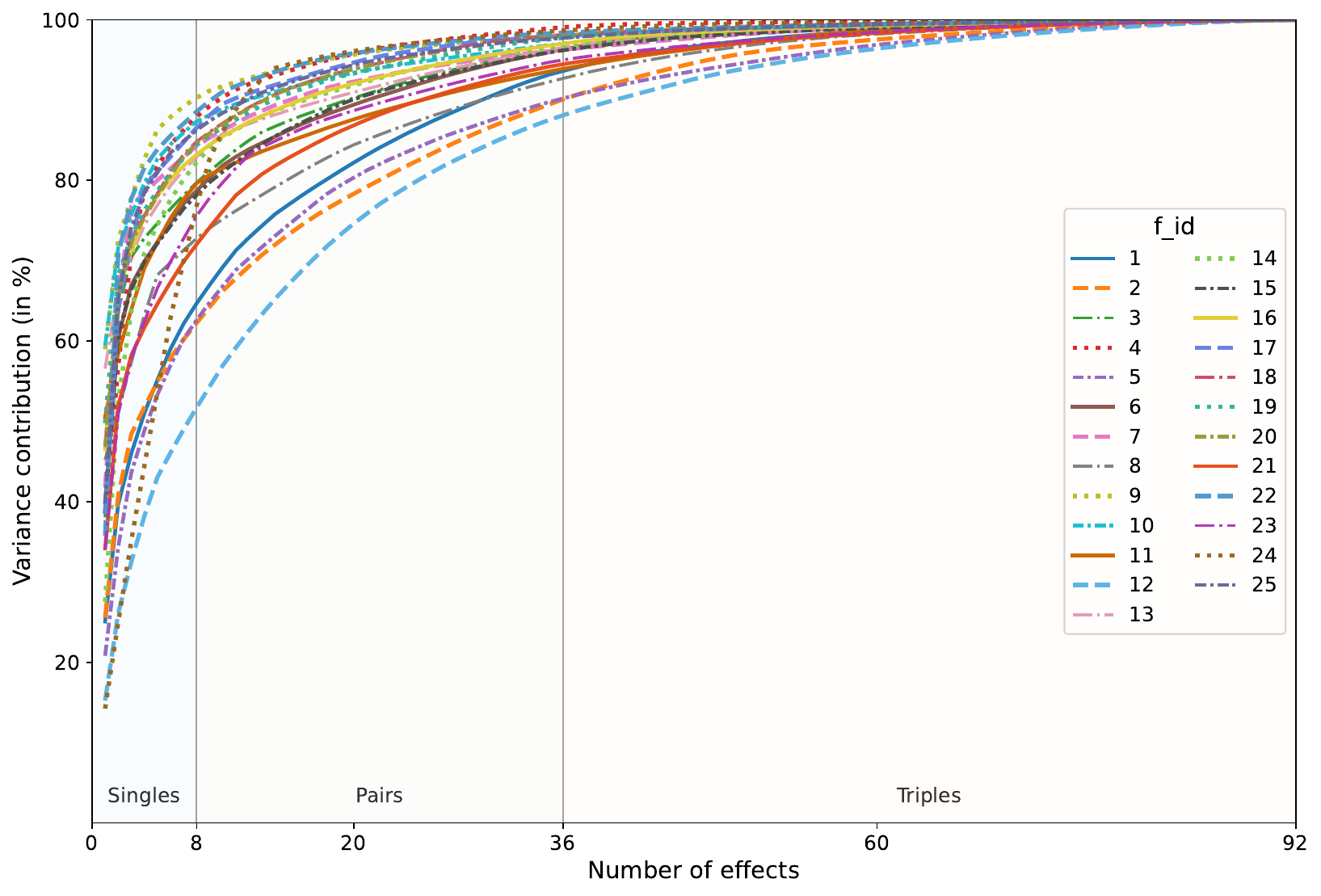}\label{fig:cumvar30}}
    \caption{Number of effects vs. cumulative variance contribution of the 92 module effects on each of the 25 problem classes belonging to the CEC'05 suite.}
\end{figure*}

\Figs~\ref{fig:cumvar10} and \ref{fig:cumvar30} show the cumulative variance contribution of ranked modules effects in the \PSOX framework for the 10$D$ and 30$D$ problems, respectively. 
The $x$-axis denotes the number of included module effects (from the most influential up to the 92nd), while the $y$-axis indicates the cumulative percentage of performance variance explained by the effects.
Each curve corresponds to one of the 25 CEC’05 benchmark functions, and different segments of the curves reflect contributions from first-order (single module) effects, second-order (pairwise) interaction effects, and third-order (triples) interaction effects. In these plots, we observe that a large portion of performance variance is usually captured by a relatively small subset of top-ranked effects, although the rate of this accumulation varies considerably across the functions.

For the majority of problems, the initial slope of the cumulative importance curves is steep, indicating that the effect of the single modules account for a substantial share of the variance. 
For example, in both the 10$D$ and 30$D$ plots, several functions exhibit curves that rise rapidly toward 50–60\% variance explained within the first 5 effects.
This suggests that, for these problems, performance differences are dominated by a few of influential modules and their interaction. 
In the 10$D$ results (\Fig\ref{fig:cumvar10}), functions such as $f_{7}$ and $f_{9}$ and $f_{10}$ exemplify this behavior. Their curves surge to a high percentage (over 60–70\% of total variance) within roughly 5–10 top-ranked effects, after which additional effects yield diminishing returns. 
In the 30$D$ results (\Fig\ref{fig:cumvar30}), we see a similar trend for functions $f_{7}$ and $f_{9}$ and $f_{10}$, but the curves are slightly less steep than in 10$D$.
Regardless of dimensionality, there is a high cumulative variance with relatively few effects. This suggests that most functions do not induce extreme module interaction response. In \Sect\ref{sec:clustering}, we identify the key modules used by \PSOX for each problem class and explain how and why they contribute to its performance.

Despite the common trend of a few module effects being highly influential, we also note clear differences among functions.
A small subset of functions across the entire benchmark show curves that have a more gradual ascent.
For these functions, no single module or pair of modules dominates the performance; instead, variance accumulates slowly as many effects are added. 
For example, in the 10$D$ result, functions in the middle and latter part of the benchmark (e.g. multimodal function $f_{12}$ and composite functions $f_{24}$) have nearly linear growth curves---even the first 10 effect explain only half of the variance. 
In the 30$D$ results, the curve for multimodal function $f_{12}$ shows similar behavior, but unimodal functions $f_{1}$-$f_{2}$ and hybrid composition function $f_{25}$ also remain relatively low within the first dozens of effects.
This indicates that the performance on these functions depends on a broad combination of modules and their interactions. 

By comparing the 10$D$ and 30$D$ results, we observe that increasing the problem dimension tends to slightly flatten the importance curves overall. 
Indeed, while the rank ordering of effects is similar, the results in \Fig\ref{fig:cumvar30} show a more gradual accumulation of variance than in \Fig\ref{fig:cumvar10}. In other words, higher-dimensional instances require considering more module effects to reach the same cumulative variance threshold. 
For instance, on $f_1$ the 10$D$ curve rises above 60\% within 5 effects, whereas in 30$D$ it rises a bit more slowly as the first 5 single effects cover only around 40–45\% of variance.
A similar trend holds for several multimodal functions: the contribution of first-order effects is somewhat reduced in 30$D$, implying that second- and third-order interactions contribute relatively more to performance variability in higher dimensions.
Finally, it is worth noting that the variance in performance across the entire CEC'05 suite can be explained by considering the combined effects of maximum three modules, suggesting that while higher-order interactions may exist, they have a negligible influence on the performance of \PSOX algorithms. 

\subsection{Clustering of Problem Classes by Module Effects}\label{sec:clustering}



\begin{figure*}[t]
    \centering
    \subfloat[10$D$ problems.]{
        \includegraphics[width=0.45\textwidth]{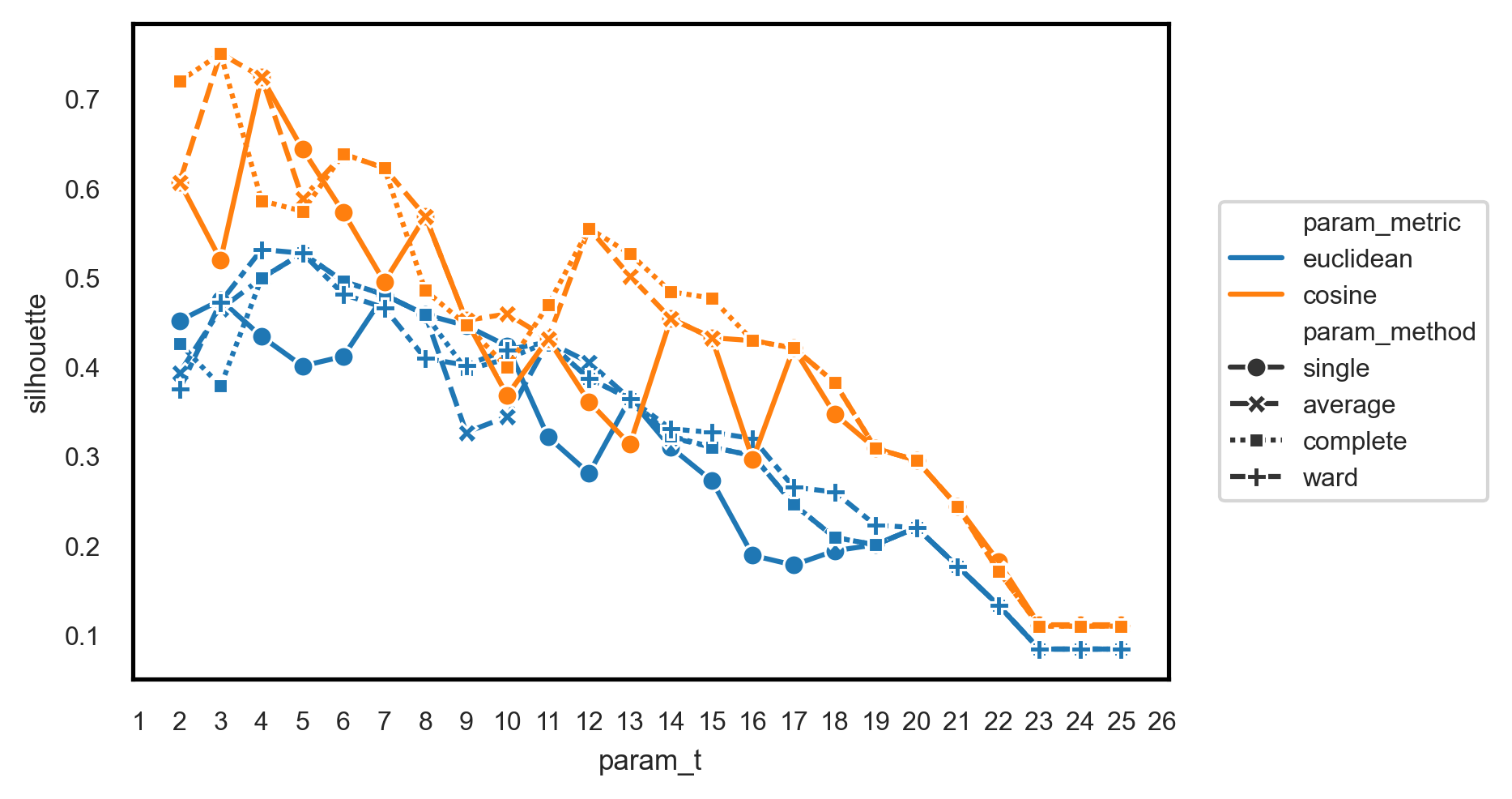}\label{fig:silhouette10}}
    \quad
    \subfloat[30$D$ problems.]{
        \includegraphics[width=0.45\textwidth]{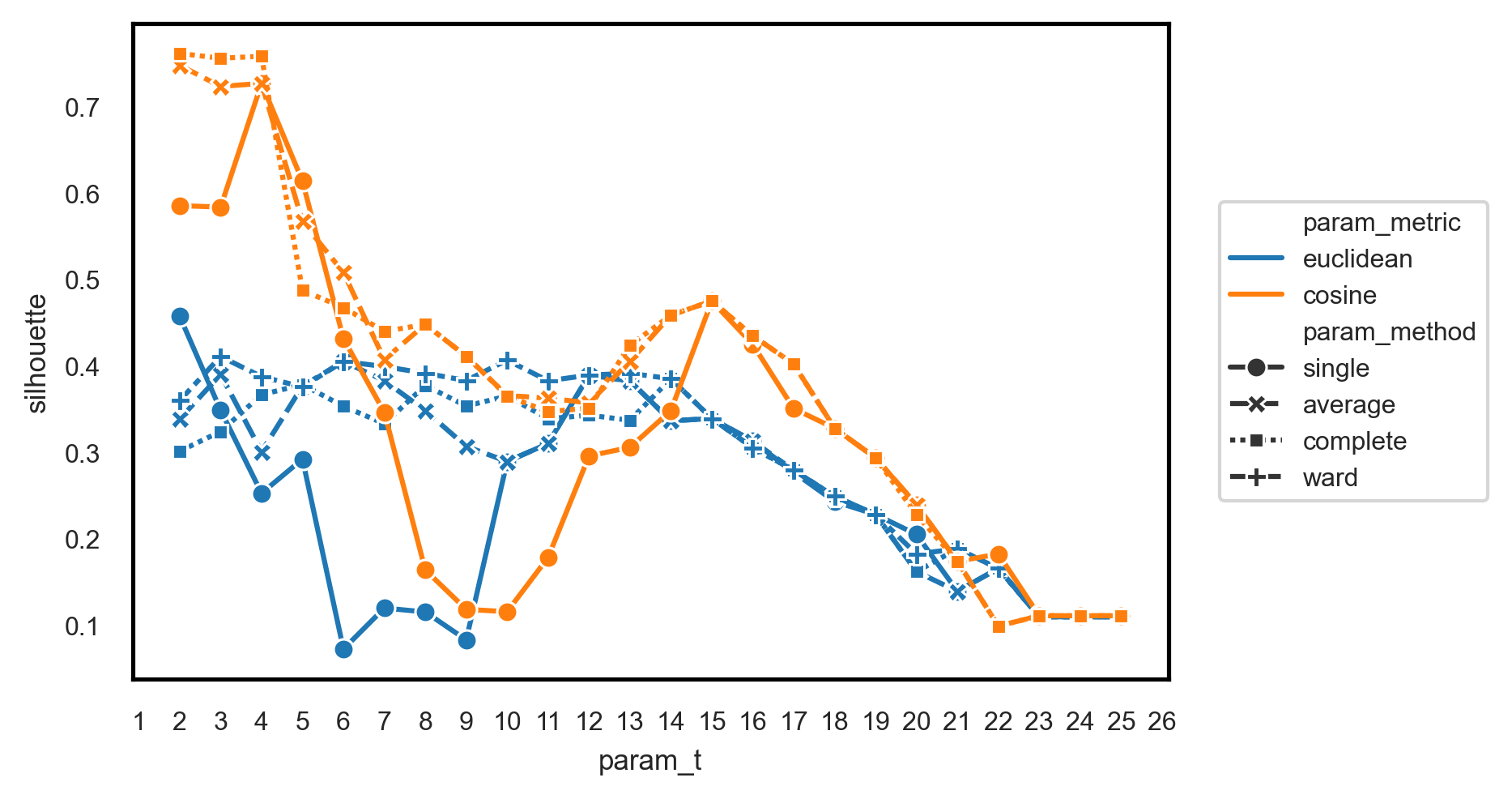}\label{fig:silhouette30}}
    \caption{Performance results from the tuning of the HC algorithm. Silhouette coefficient is used as clustering quality measure.}
    \label{fig:clustering_silhouette}
\end{figure*}

We applied hierarchical clustering (HC) to the 25-dimensional module-effect vectors obtained via \fANV in order to group problem classes with similar module importance patterns. 
Before clustering, we tuned the HC algorithm’s hyperparameters (number of clusters ($k$), linkage method, and distance metric) by evaluating clustering quality under various settings. 
\Fig\ref{fig:clustering_silhouette} display the Silhouette coefficient for (i) varying values of $k$, from 2 to 25, (ii) two distance metrics (\textit{Euclidean} and \textit{cosine}), and (iii) four linkage methods (\textit{single}, \textit{complete}, \textit{average}, \textit{ward}). 
Each line in \Figs\ref{fig:silhouette10} (10$D$ problems) \ref{fig:silhouette30} (30$D$ problems) represent one combination of distance metric and linkage. 

\begin{figure*}[t]
    \centering
    \subfloat[10$D$ problems.]{
        \includegraphics[width=0.4\textwidth]{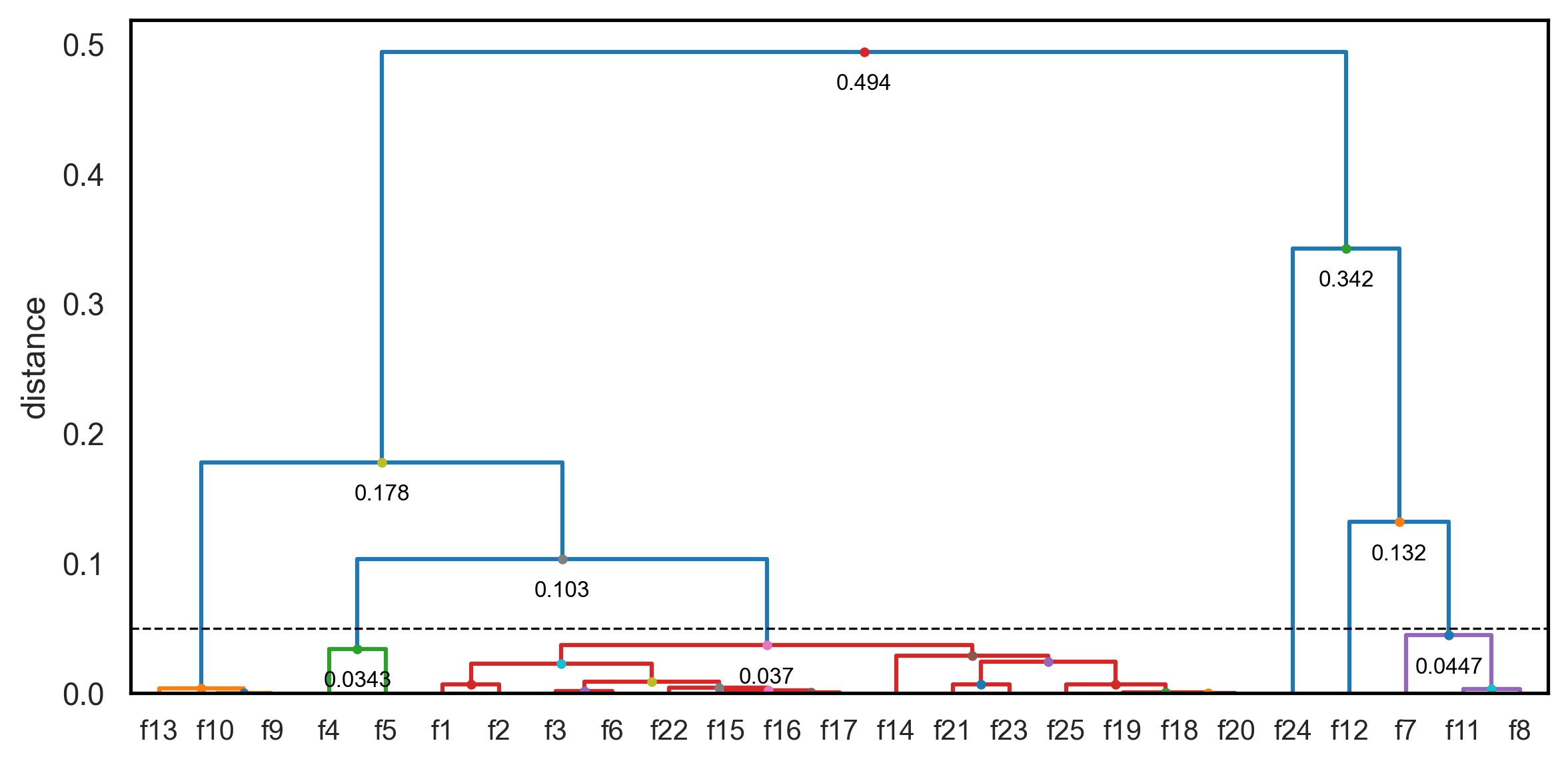}\label{fig:dendrogram10}}
    \quad
    \subfloat[30$D$ problems.]{
        \includegraphics[width=0.4\textwidth]{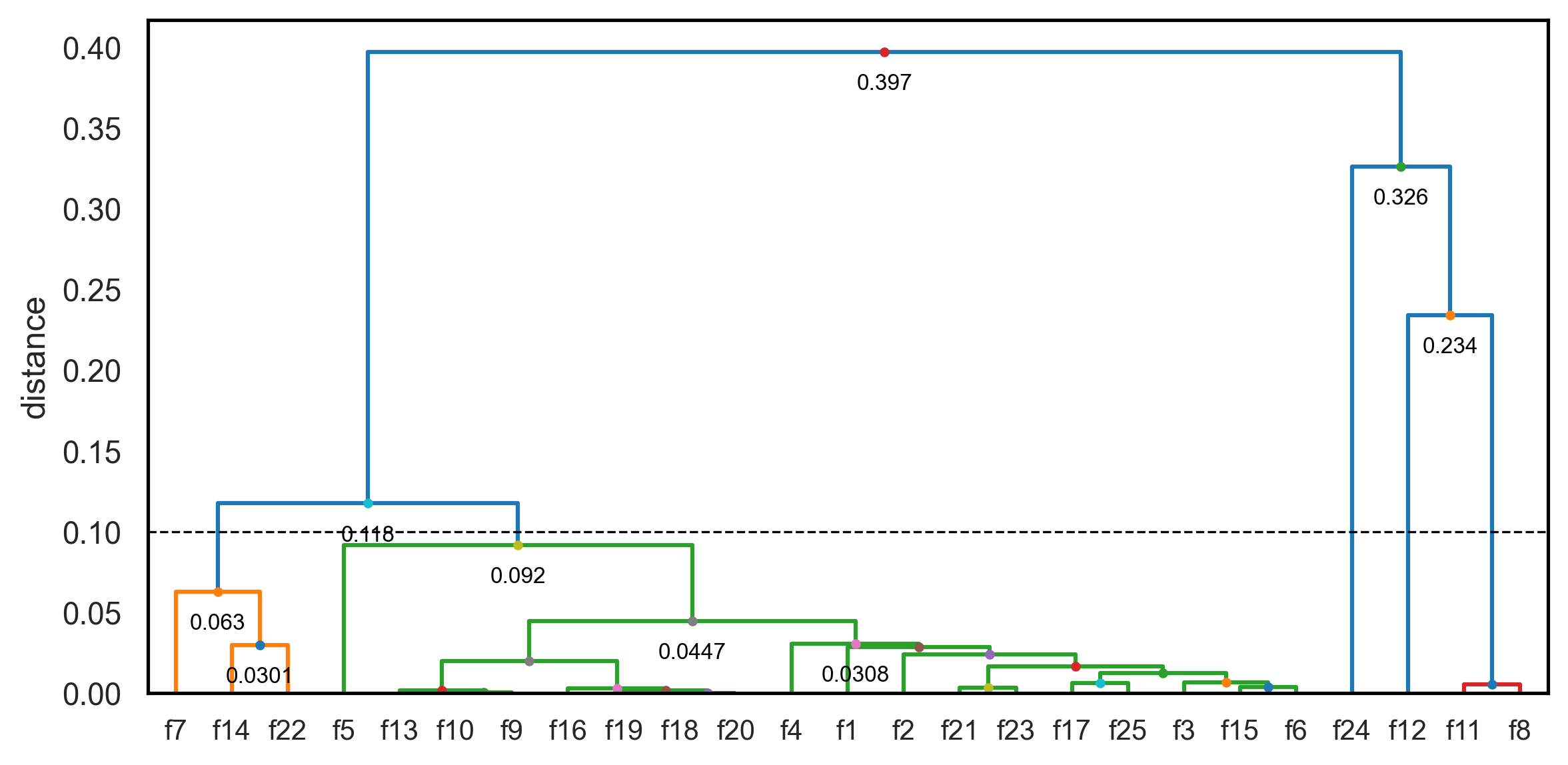}\label{fig:dendrogram30}}
    \caption{Dendrogram of the clustering process of HC. It visualizes how problem classes are iteratively grouped based on similarity, with branch heights indicating inter-cluster dissimilarity. Color coding highlights the resulting clusters.}
    \label{fig:clustering_dendrogram}
\end{figure*}

We found that a moderate number of clusters---between 5 and 7---yielded the highest Silhouette scores, with \textit{cosine} distance and \textit{complete} linkage performing best overall. Based on this, for our final clustering, we selected $k=6$ clusters for the 10$D$ problems, $k=5$ clusters for the 30$D$ problems, and \textit{cosine} distance and \textit{complete} linkage for both 10$D$ and 30$D$ problems. 
The Silhouette score under these settings is about $0.6$, indicating moderately well-defined clusters; however, the relatively low maximum inter-cluster distance (approximately $0.065$ for 10$D$ problems and $0.1$ for 30$D$ problems on the normalized variance scale) suggests that all problem classes share broadly similar module importance profiles. This means that, even though distinct clusters can be identified, the differences between clusters are subtle and the overall patterns of module contributions do not vary drastically across the benchmark.

In \Fig~\ref{fig:clustering_dendrogram}, we show dendrograms, which are a tree-like diagram representing the hierarchical clustering process, for each of the 10$D$ and 30$D$ problems. 
The vertical axis indicates inter-cluster distance at each merge. The dashed line shows the chosen cut yielding $6$ clusters for the 10$D$ results (\Fig\ref{fig:dendrogram10}) and $5$ clusters for the 30$D$ results (\Fig\ref{fig:dendrogram30}).
The red dots in the branches mark cluster merge distances.
The highest merge distance (0.494, in \Fig\ref{fig:dendrogram10}, and 0.397, in \Fig\ref{fig:dendrogram30}) and the relatively low overall distances confirm that module importance patterns are generally similar across classes.
In the following, we analyze the grouping of problem classes based on the module effects using the results from the HC process.

\begin{figure*}[thb]
    \centering
   \includegraphics[width=1.0\textwidth]{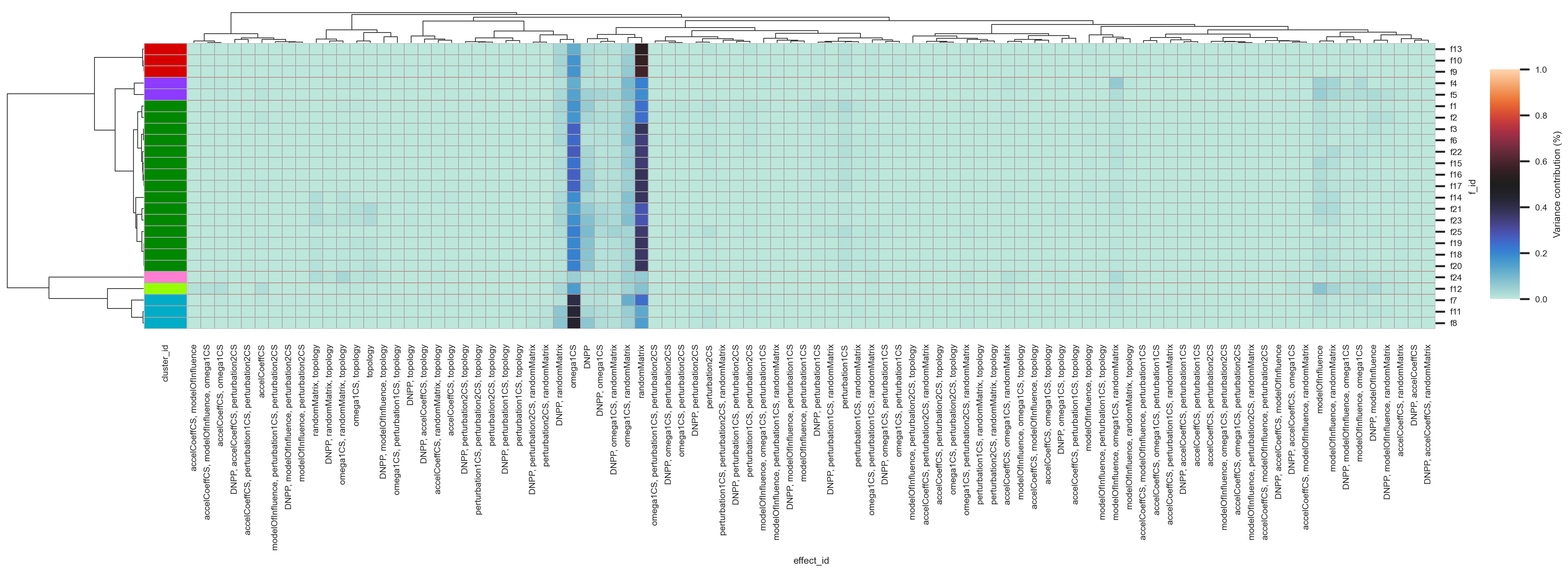}
    \caption{Clustermap of the 10$D$ problems.}
    \label{fig:clustermap10}
\end{figure*}

\subsubsection{Results on the 10$D$ Problems}\label{sec:Results on the 10D Problems}
Figure~\ref{fig:clustermap10} depicts a clustermap of the vector representation of the 10$D$ problems. Rows correspond to problem classes ($f_{\text{id}}$), and columns to module effects (\textit{effect}$_{\text{id}}$). The color intensity indicates the variance contribution of each module effect, ranging from 0 (no importance) to 1 (high importance). The first column shows cluster assignments, with each color representing a different cluster. The results reveal that most problem classes fall into three dominant clusters ({\texttt{dark green}}, \texttt{red} and \texttt{cyan}), while a few remain isolated in smaller, distinct clusters (\texttt{purple}, \texttt{pink} and \texttt{light green}).
In the clustermap, we can see that the most influential modules (i.e., those that produce stronger individual effects) across all problem classes are the \rndMtx and \OmgCS modules, and to a lesser extent, the \DNPP module. 
The \Moi and \PertInfCS modules are also activated for some functions, but their contribution, individual and combined, is small.
Interestingly, the clustermap also shows that the individual effect of the \Topo, \AcCof and \PertRndCS, and the vast majority of their pairwise and triple interactions do not produce any meaningful effect on the performance of \PSOX. 
Among the interaction effects, combinations involving \OmgCS $+$ \rndMtx, \DNPP $+$ \OmgCS, \DNPP $+$ \rndMtx, and the triplet \DNPP $+$ \OmgCS $+$ \rndMtx stand out as highly impactful. This suggests that the behavior of the \PSOX variants is primarily driven by the interplay of these three modules.

\begin{figure*}[thb]
    \centering
    \makebox[\textwidth][c]{%
        {\includegraphics[width=0.15\textwidth]{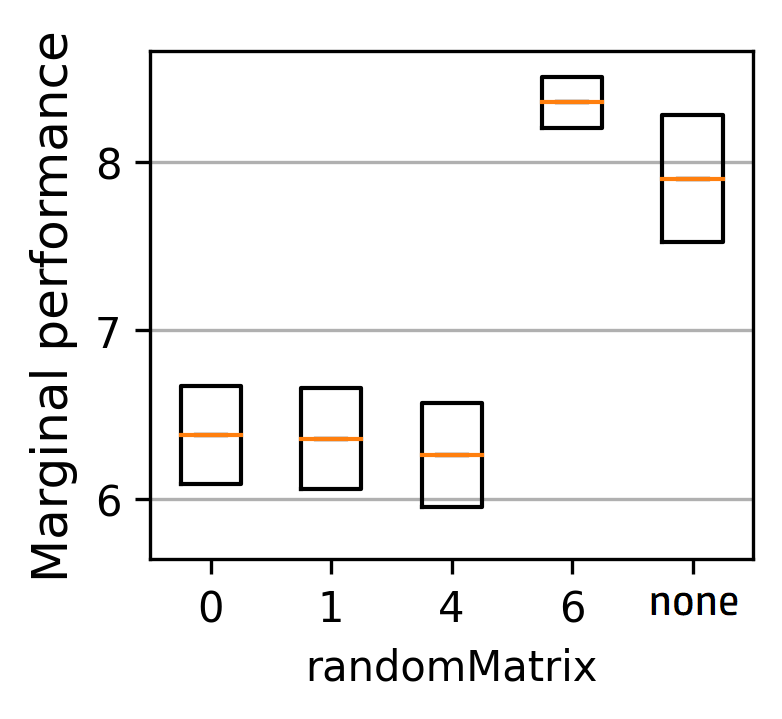}}\quad
        {\includegraphics[width=0.15\textwidth]{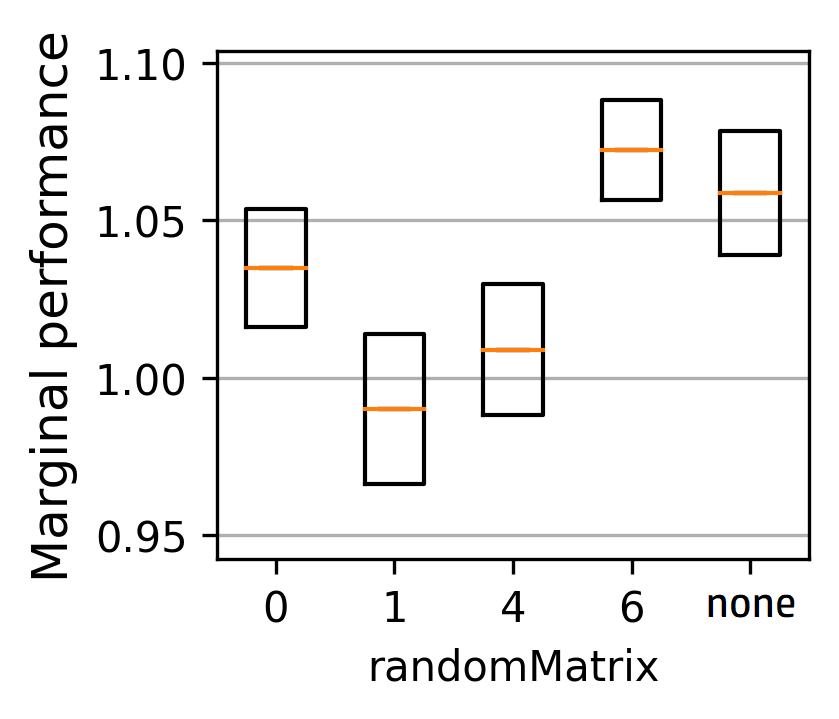}}\quad
        {\includegraphics[width=0.15\textwidth]{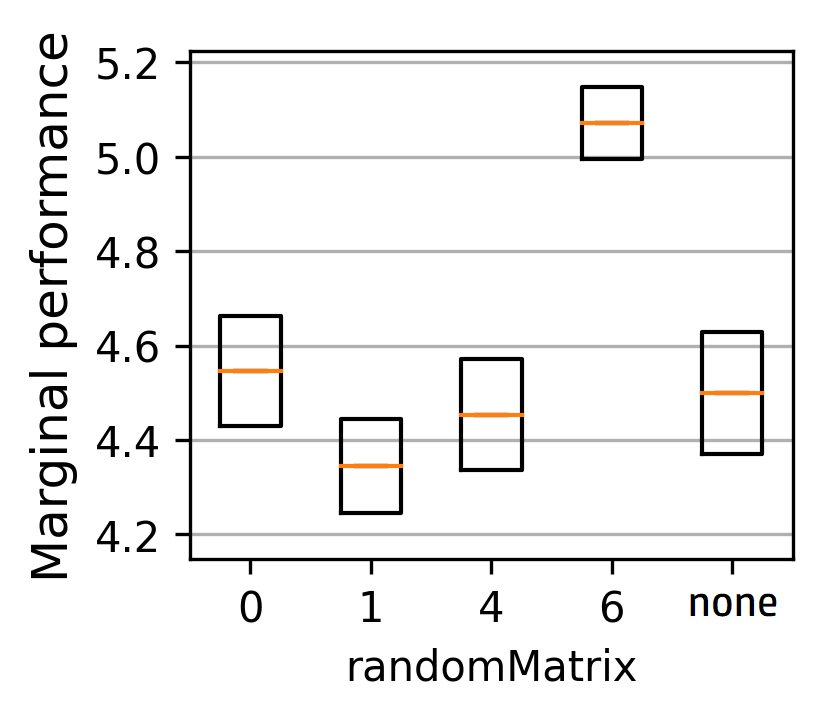}}\quad
        {\includegraphics[width=0.15\textwidth]{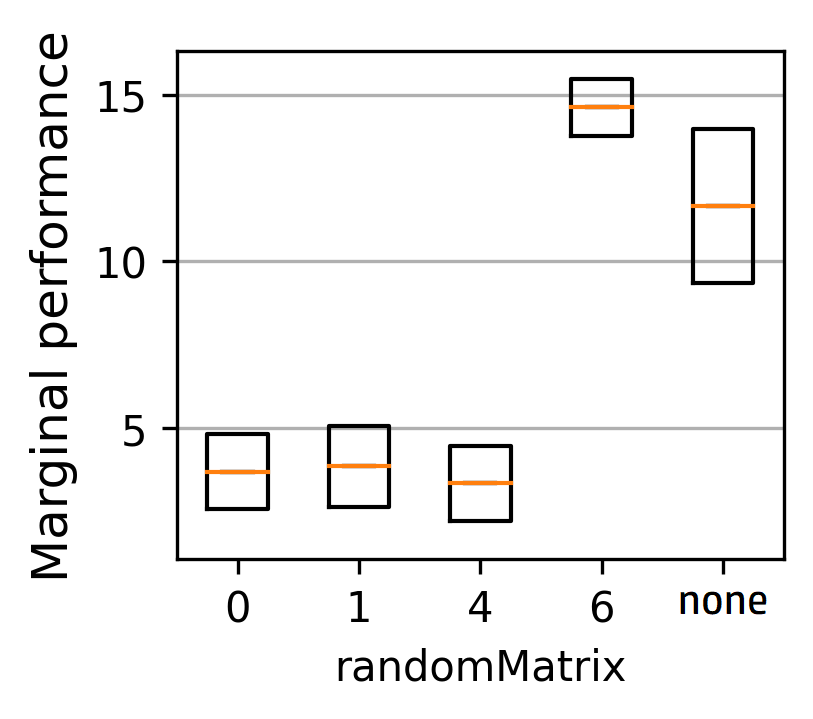}}\quad
        {\includegraphics[width=0.15\textwidth]{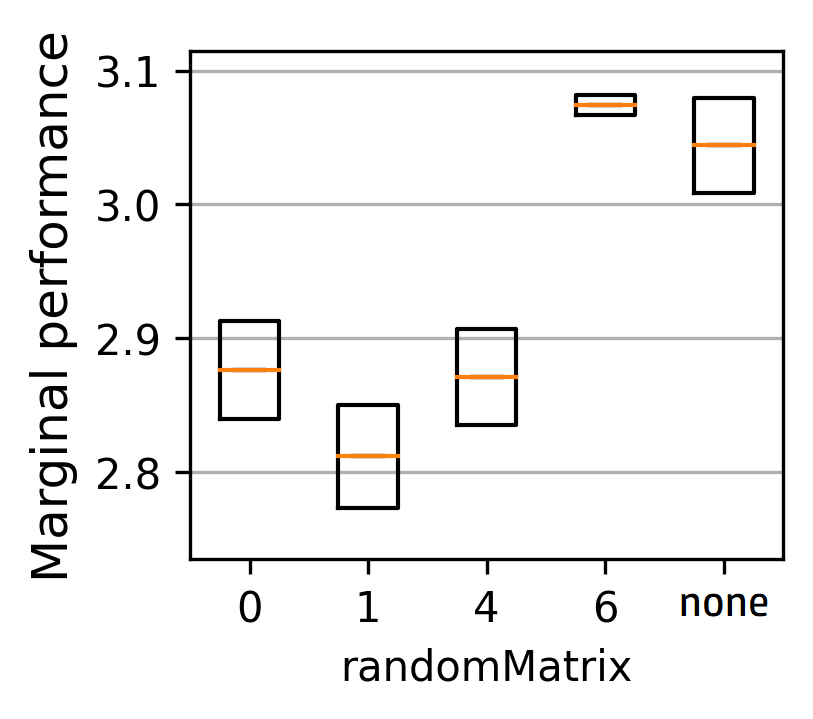}}\quad
        {\includegraphics[width=0.15\textwidth]{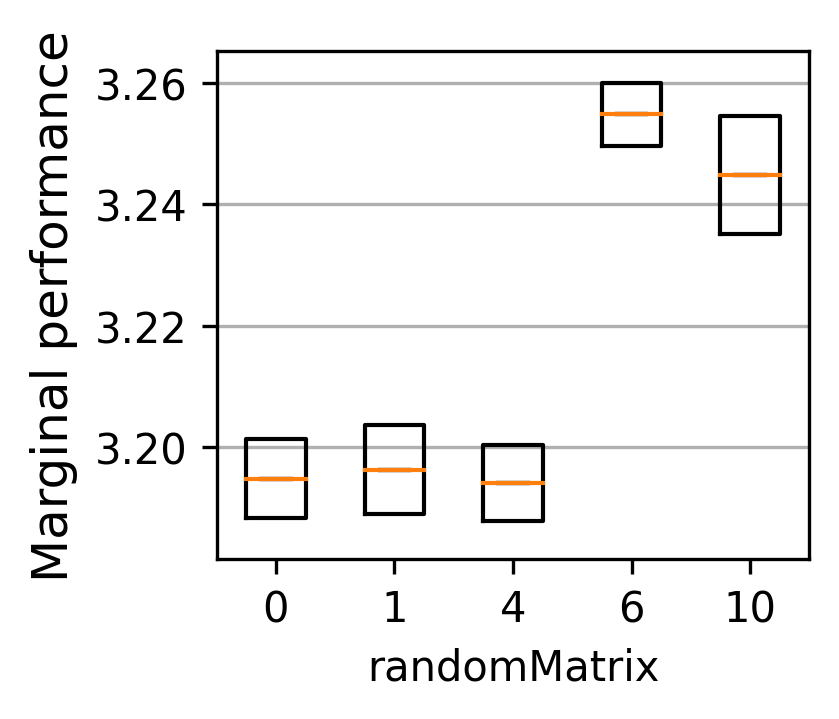}}
        }
    \makebox[\textwidth][c]{%
        {\includegraphics[width=0.15\textwidth]{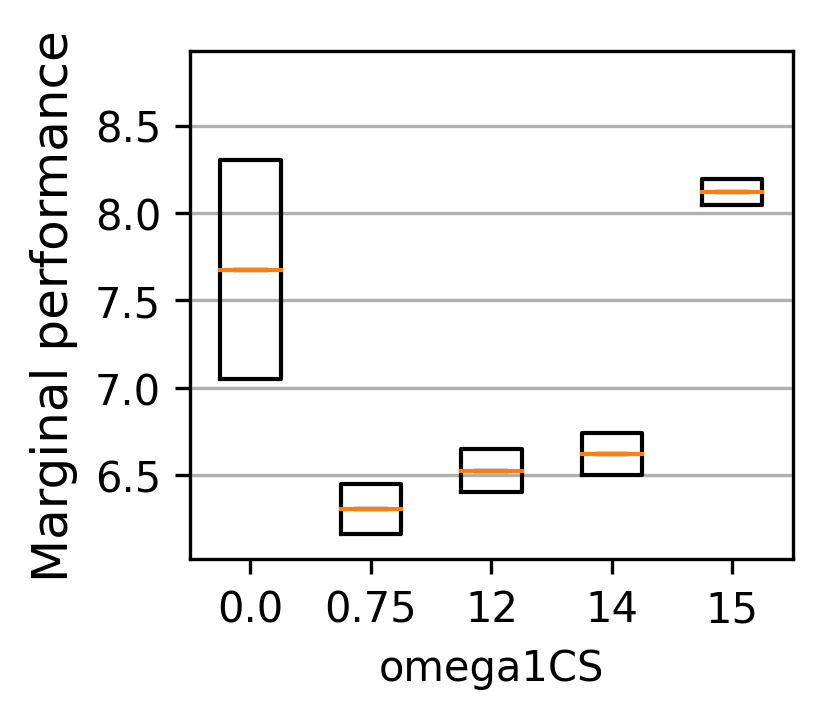}}\quad
        {\includegraphics[width=0.15\textwidth]{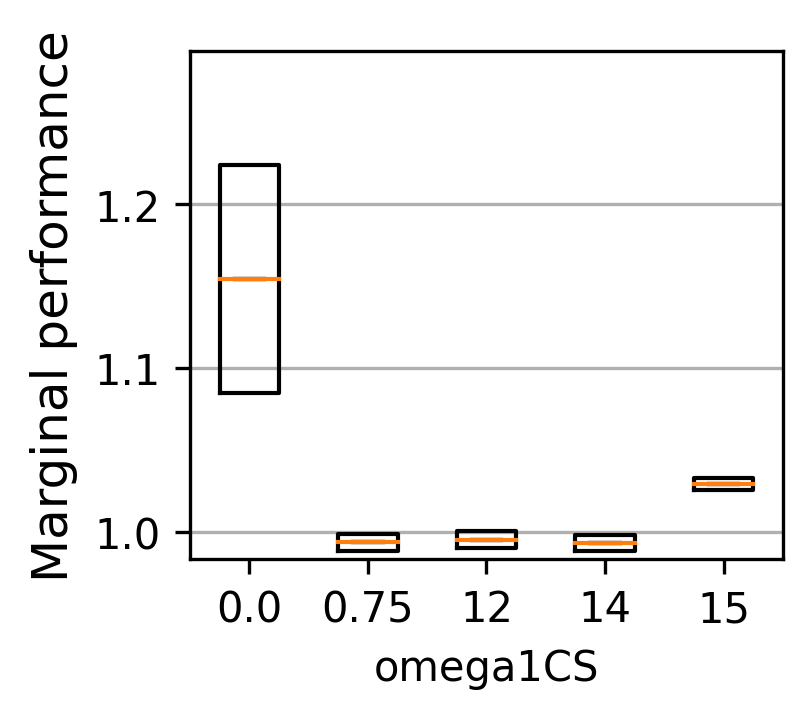}}\quad
        {\includegraphics[width=0.15\textwidth]{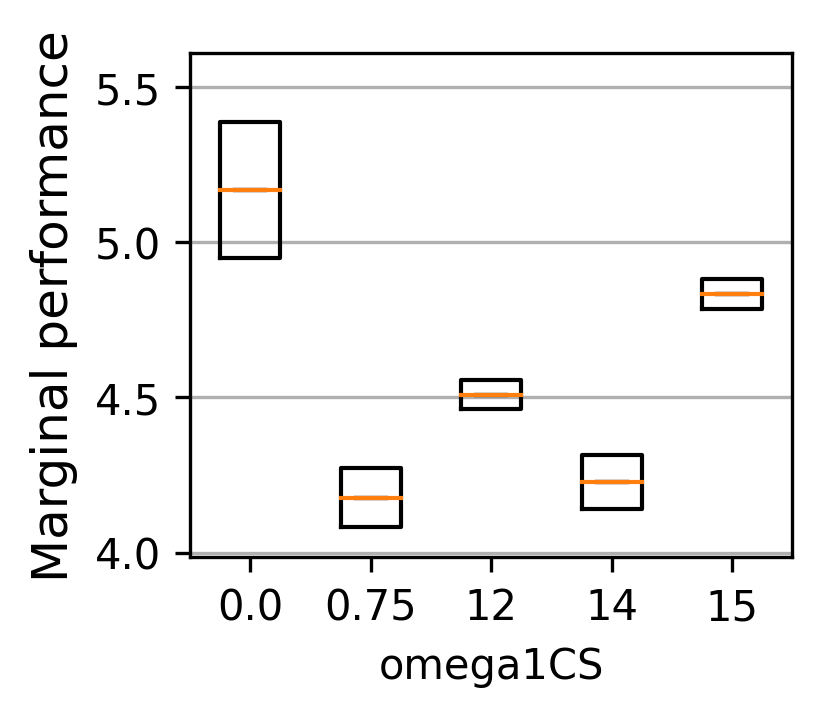}}\quad
        {\includegraphics[width=0.15\textwidth]{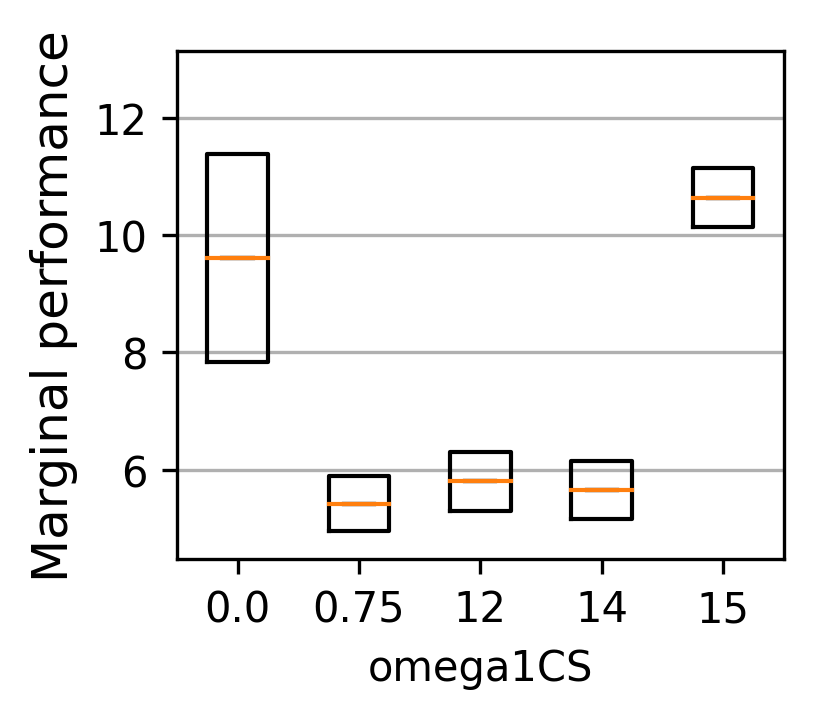}}\quad
        {\includegraphics[width=0.15\textwidth]{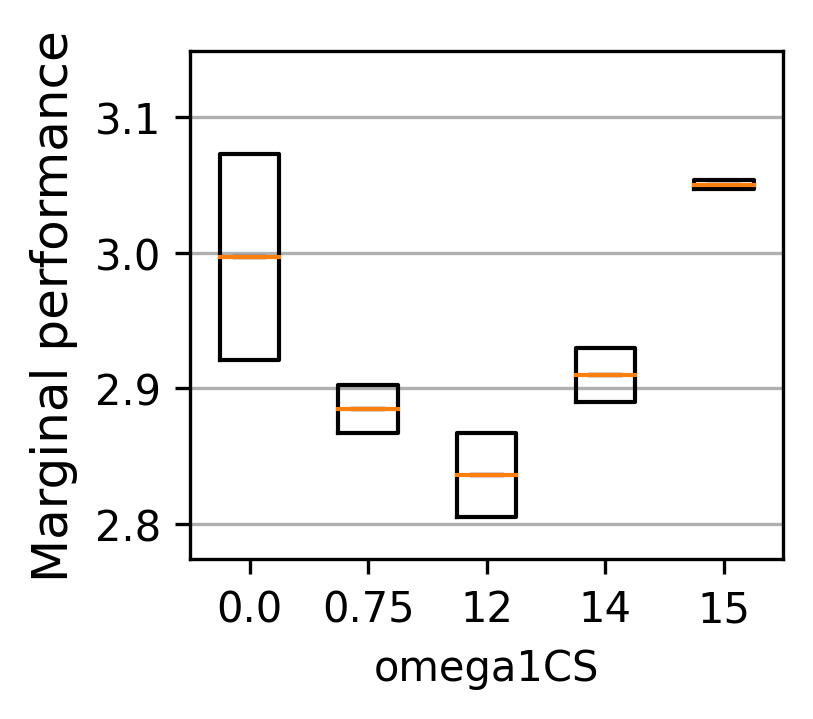}}\quad
        {\includegraphics[width=0.15\textwidth]{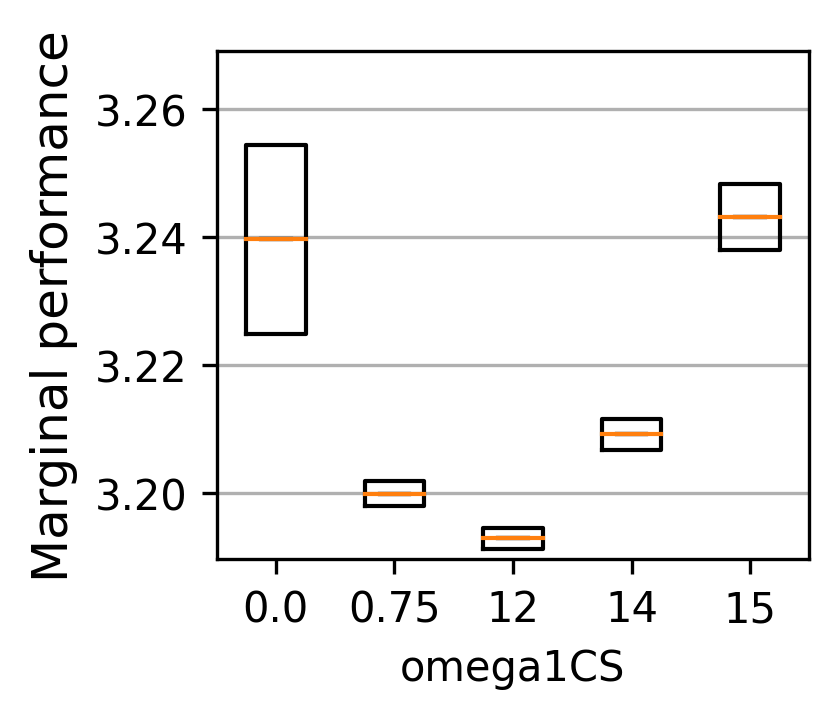}}
    }
    \makebox[\textwidth][c]{%
        \subfloat[$f_{3}$]
        {\includegraphics[width=0.15\textwidth]{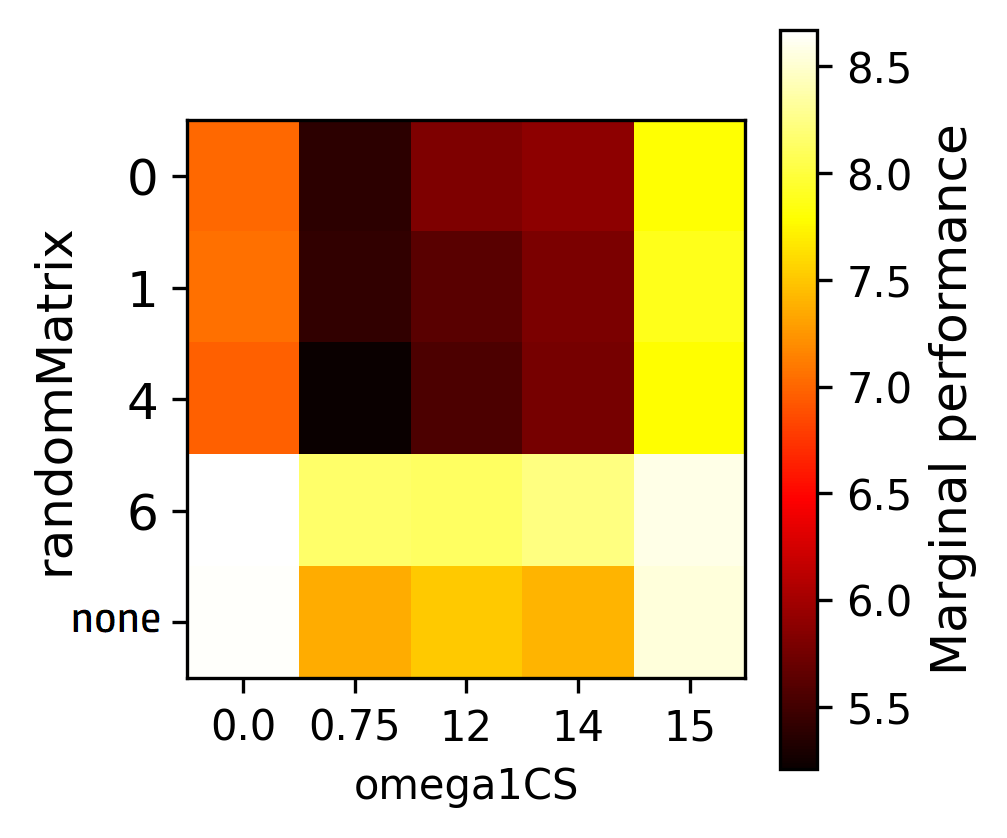}\label{fig:omega1CS_randomMatrix_2_10D}}\quad
        \subfloat[$f_{11}$]
        {\includegraphics[width=0.15\textwidth]{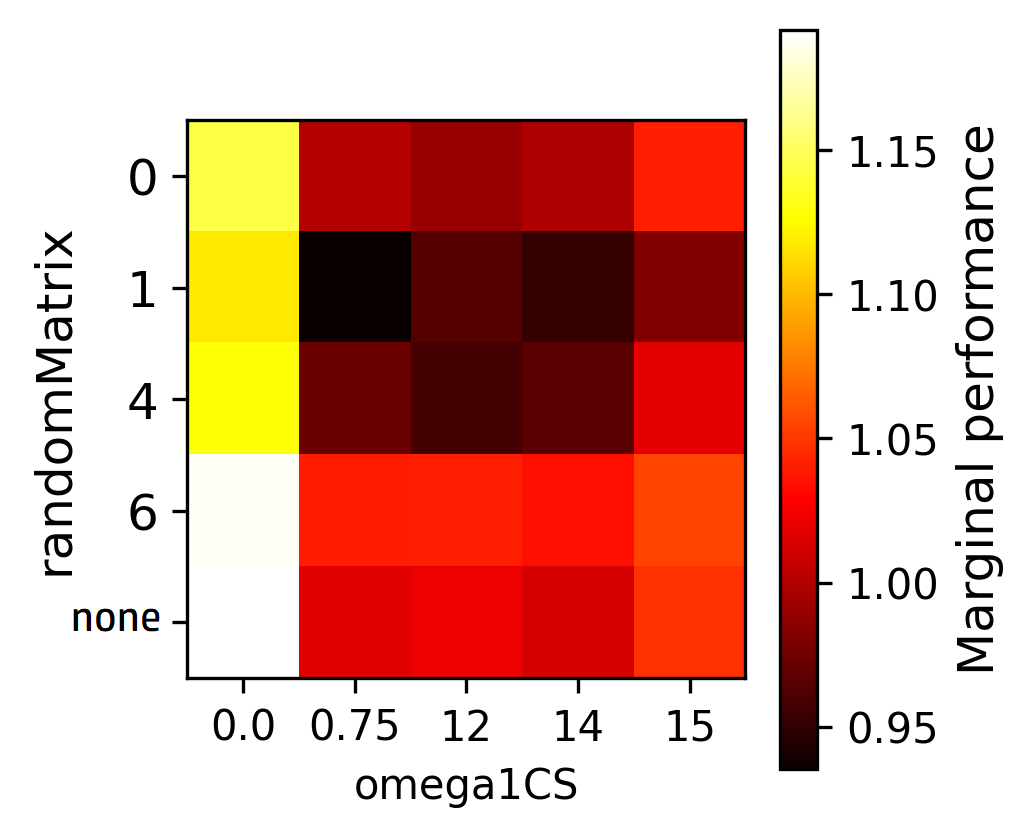}\label{fig:omega1CS_randomMatrix_10_10D}}\quad
        \subfloat[$f_{12}$]
        {\includegraphics[width=0.15\textwidth]{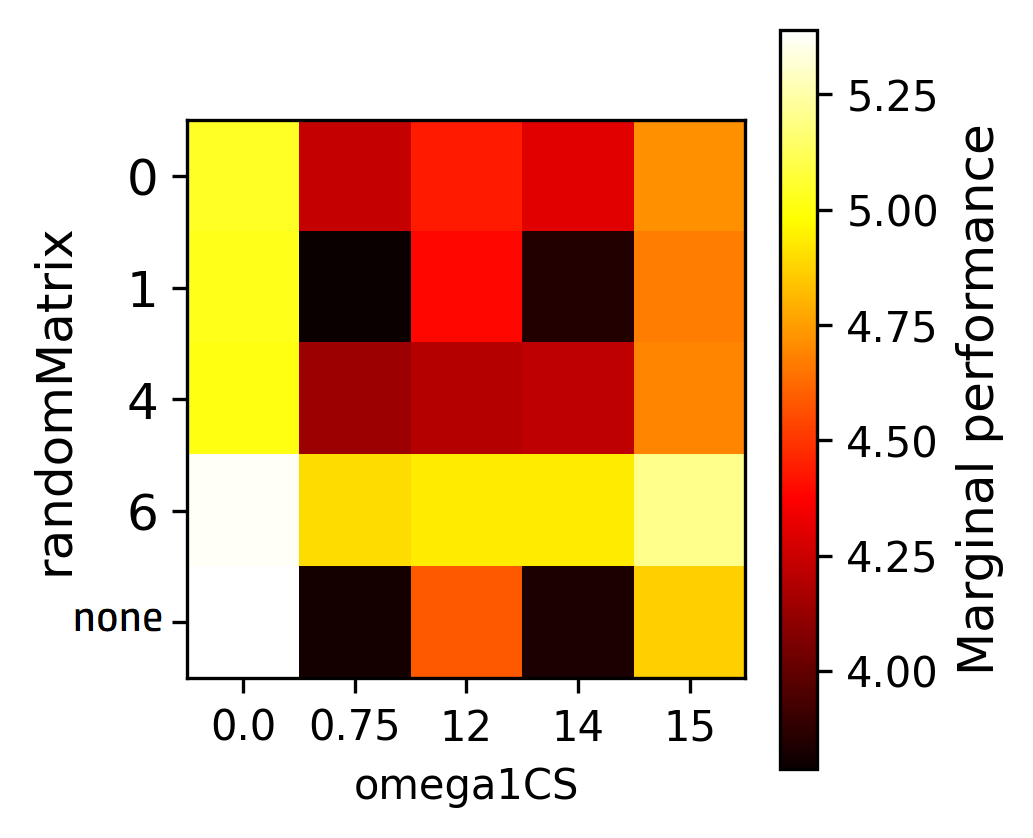}\label{fig:omega1CS_randomMatrix_11_10D}}\quad
        \subfloat[$f_{13}$]
        {\includegraphics[width=0.15\textwidth]{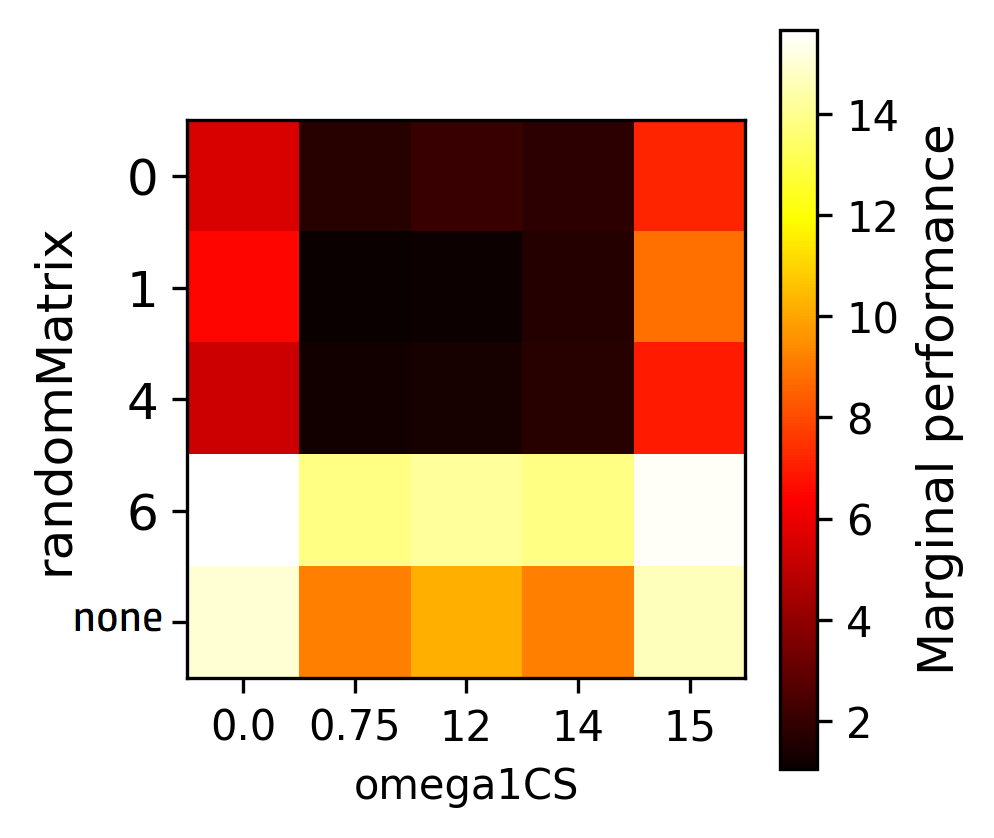}\label{fig:omega1CS_randomMatrix_12_10D}}\quad
        \subfloat[$f_{23}$]
        {\includegraphics[width=0.15\textwidth]{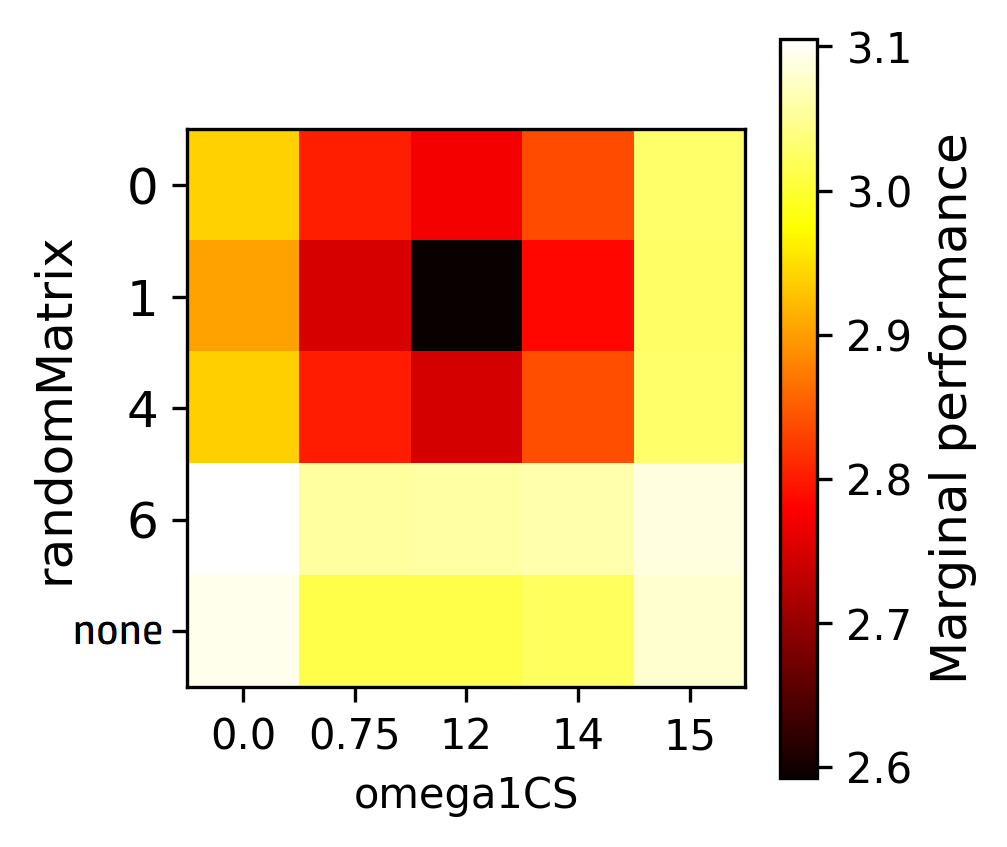}\label{fig:omega1CS_randomMatrix_22_10D}}\quad
        \subfloat[$f_{24}$]
        {\includegraphics[width=0.15\textwidth]{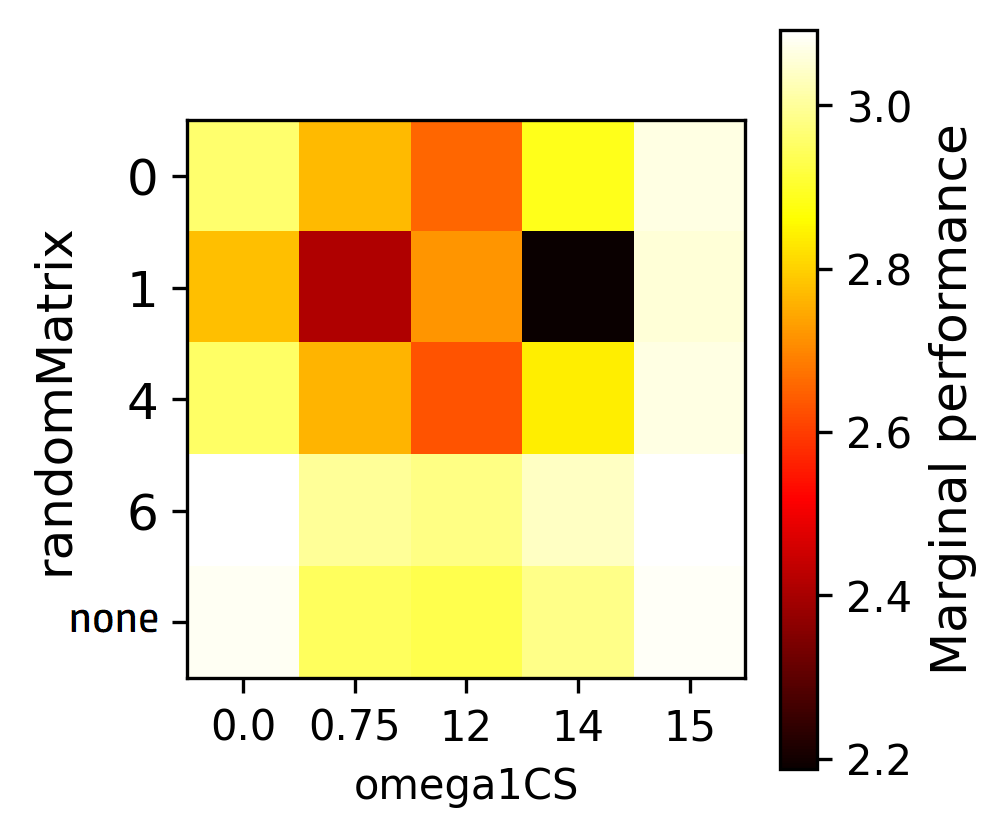}\label{fig:omega1CS_randomMatrix_24_10D}}
    }
    \caption{Marginal performance of the options for the \rndMtx and \OmgCS modules, in problem classes 
     $f_3$ (\texttt{purple} cluster), $f_{11}$ (\texttt{cyan} cluster), $f_{12}$ (\texttt{light green} cluster), $f_{13}$ (\texttt{red} cluster), $f_{23}$ (\texttt{dark green} cluster) and $f_{24}$ (\texttt{pink} cluster)
     with $D=10$.}
    \label{fig:marginal_performance_10D}
\end{figure*}

\Fig\ref{fig:marginal_performance_10D} depicts the marginal performance of the implementation options defined for the \rndMtx and \OmgCS modules in 
representative examples of functions belonging to different clusters, namely \texttt{purple} ($f_3$), \texttt{cyan} ($f_{11}$), \texttt{light green} ($f_{12}$), \texttt{red} ($f_{13}$), \texttt{dark green} ($f_{23}$) and \texttt{pink} ($f_{24}$).
The first two upper rows of \Fig\ref{fig:marginal_performance_10D} display the marginals for the individual effect as boxplots, the $x$-axis shows the different options for the module, and the $y$-axis indicates the marginal performance, where lower values indicate better marginal performance (i.e., lower distance to the optimum on average). The heatmaps in the bottom row of \Fig\ref{fig:marginal_performance_10D} depict the marginal performance for a combination of module options, the $x$- and $y$-axis show the different implementation options for the modules, while the color-coding indicates the marginal performance, where lower values indicate better performance. 

The \rndMtx and \OmgCS modules define how particles move in the search space. Thus, their impact on \PSOX performance is particularly noticeable in problems involving rotations of the search space and multiple local optima. 
In practice, the \rndMtx module provides options to generate random transformation matrices that rotate and/or scale the velocity vector of the particles ($\vec{v}^{i}_{t+1}$), whereas \OmgCS provide options to control particles' inertia ($\omega_1$) in the generalized velocity update rule---see \Tbl\ref{tab:psox_components}.
Combining these two modules enables particles to explore different hyperplanes in the search space and adjust their step size from very long to very small displacements.
The implementation options that consistently yield the best marginal performance across all problem classes are options $0=\MtxIdentity$, $1=\MtxDiagonal$ and $4=\MtxEuclidean$, for the \rndMtx module; and $0=\OmegaCons$, $12=\OmegaAdapVel$ and $14=\OmegaRnkBsd$, for the \OmgCS module.
Conversely, the worst options for the \OmgCS module seem to be the two extremes: 0=\OmegaCons ($\omega_1 = 0.0$), which eliminates the inertia term, and 15=\OmegaSuccBsd, which leads to rapid adaptation of the inertia term once improvements are observed leading to premature convergence.

As depicted in \Fig\ref{fig:marginal_performance_10D}, the functions in the \texttt{red} cluster ($f_9$, $f_{10}$, and $f_{13}$) form a compact group with consistently high marginal importance of \rndMtx, followed by \OmgCS, and their interaction by a smaller extent. 
Among the 25 functions in the CEC'05 benchmark, functions $f_9$, $f_{10}$, and $f_{13}$ are the ones where the marginal performance is most strongly influenced by the \rndMtx module. 
All three function in this cluster have similar bowl-like shape and share pronounced multimodality, relative to other multimodal functions (e.g., Rosenbrock function, $f_6$) or the more "easy" unimodal set ($f_{1}$-$f_5$).
Also, function $f_{10}$ is the rotated version of function $f_9$, both with regular repeating local optima.
These patterns suggest that the performance of \PSOX on this class of problems depends largely on the search basis. Different rotation bases allow particles to observe the search space in different hyperplanes, which allows them to distinguish more easily among valleys with different quality. Additionally, the performance on \PSOX seems to depend also on how quickly inertia is reduced, with a focus on exploration in the early stages of the optimization process and exploitation thereafter.

The \texttt{cyan} cluster, composed of functions $f_7$, $f_8$, and $f_{11}$, is on the opposite end of the spectrum in terms of how the influence of \rndMtx and \OmgCS is balanced out.
The three functions in the \texttt{cyan} cluster are the ones where the effect of the \OmgCS is the strongest, while the effect of the \rndMtx module becomes less relevant.
This cluster encompasses functions with challenging global structures, such as optima outside the initialization range, boundary optima, or high ruggedness, indicating a consistent algorithmic behavior across these scenarios.
The options for the \OmgCS module provides options for inertia step control and boundary handling by limiting the size of the velocity vector, $\vec{v}^{i}_{t+1}$.
A large $\vec{v}^{i}_{t+1}$ allows particles to swiftly move from one side of the search space to another, while a small $\vec{v}^{i}_{t+1}$ lets particles make small changes to their displacement in order to locate inaccessible global optima. 
With the pervasive small-scale ruggedness/noise and unbounded search spaces in functions $f_7$, $f_8$, and $f_{11}$, the options for the \OmgCS that allow particles to modulate their velocity adaptively depending on the characteristics of the landscape become paramount. 

The largest cluster is the \texttt{dark green} one, which comprises functions $f_{1}$-$f_{3}$, $f_{6}$, $f_{14}$-$f_{23}$ and $f_{25}$.
Although these functions have a diverse set characteristics, as shown in \Tbl\ref{tbl:CEC05}, we can differentiate between two distinctive features, bowl-shaped and elliptic landscapes ($f_{1}$-$f_{3}$, $f_{6}$, and $f_{14}$) and noisy landscapes ($f_{15}$-$f_{23}$ and $f_{25}$).
The \texttt{dark green} cluster is the only one that includes all three classes of functions, unimodal, multimodal and hybrid functions, but the multimodal functions are underrepresented (only 2 out of 9 multimodal functions in the CEC'05 suite) compared to the hybrid compositions functions (10 out of 11 hybrid functions in the CEC'05 suite).
The clustermap, \Fig~\ref{fig:clustermap10}, shows a balanced contribution of the \OmgCS and the \rndMtx modules, but the latter is consistently more important.
This is not totally unexpected, since most functions in this cluster have rotated search spaces. However, the fact that most functions in this cluster are non-separable suggests that \rndMtx is also useful to deal with this feature.
It is also interesting to note that, in the \texttt{dark green} cluster (particularly in functions $f_{15}$-$f_{23}$ and $f_{25}$), the \DNPP modules plays a larger role compared to the \texttt{red} and \texttt{cyan} clusters, where \DNPP contributes only marginally.
Pairwise interactions between \OmgCS and the \rndMtx are also more present that in \texttt{red} and \texttt{cyan} clusters, but they are not extreme. 

The functions in which we observe broader module interactions are the ones in the 
\texttt{purple} (functions $f_4$ and $f_5$), \texttt{light green} (function $f_{12}$) and \texttt{pink} (function $f_{24}$) clusters.
The two functions $f_4$ and $f_5$ in the \texttt{purple} cluster are unimodal, and the main distinctive features is that function $f_4$ has a rugged landscape due to added noise in fitness, while function $f_5$ has a smooth one. 
In addition to \OmgCS and \rndMtx, the \Moi module and its pairwise interaction $\Moi +\rndMtx$ have a significant influence in \PSOX performance; however, in function $f_4$, the triplet $\Moi + \OmgCS + \rndMtx$ is also present, suggesting that it may play role, albeit small, on rugged unimodal functions.
While our results show that impact of \Moi is minimal for most functions in the CEC'05 benchmark, the effect of \MoiBoN in functions $f_4$ and $f_5$ points to the fact that, in unimodal functions, particles can benefit from following the single-best particle, as opposed to following many individuals at the same time, as it is the case in the \MoiFI.

Function $f_{12}$, in the \texttt{light green} cluster, and function $f_{24}$, in the \texttt{pink} cluster, are the ones where \PSOX exhibits the most intricate module interactions.
In particular, in function $f_{12}$, there are several small contributions from \AcCof and \Moi, and their combine interaction with \OmgCS and \rndMtx.
Compared to the rest of multimodal functions, function $f_{12}$ is the only one with a smooth landscape and a well defined valley where the global optimum is located. Our results suggest that, when facing problems with smooth landscapes, the \AcCof and \Moi modules may impact the performance of \PSOX.
In the case of function $f_{24}$, however, there is no single module that can explain the variance in performance.
As shown in the clustermap---\Fig~\ref{fig:clustermap10}---and in cumulative variance plots---\Fig\ref{fig:cumvar10}---several modules contribute to the performance of \PSOX, but none of them is dominant; rather, the variance in performance accumulates over time by the individual and combine interactions of different modules.

\begin{figure*}[thb]
    \centering
   \includegraphics[width=1.0\textwidth]{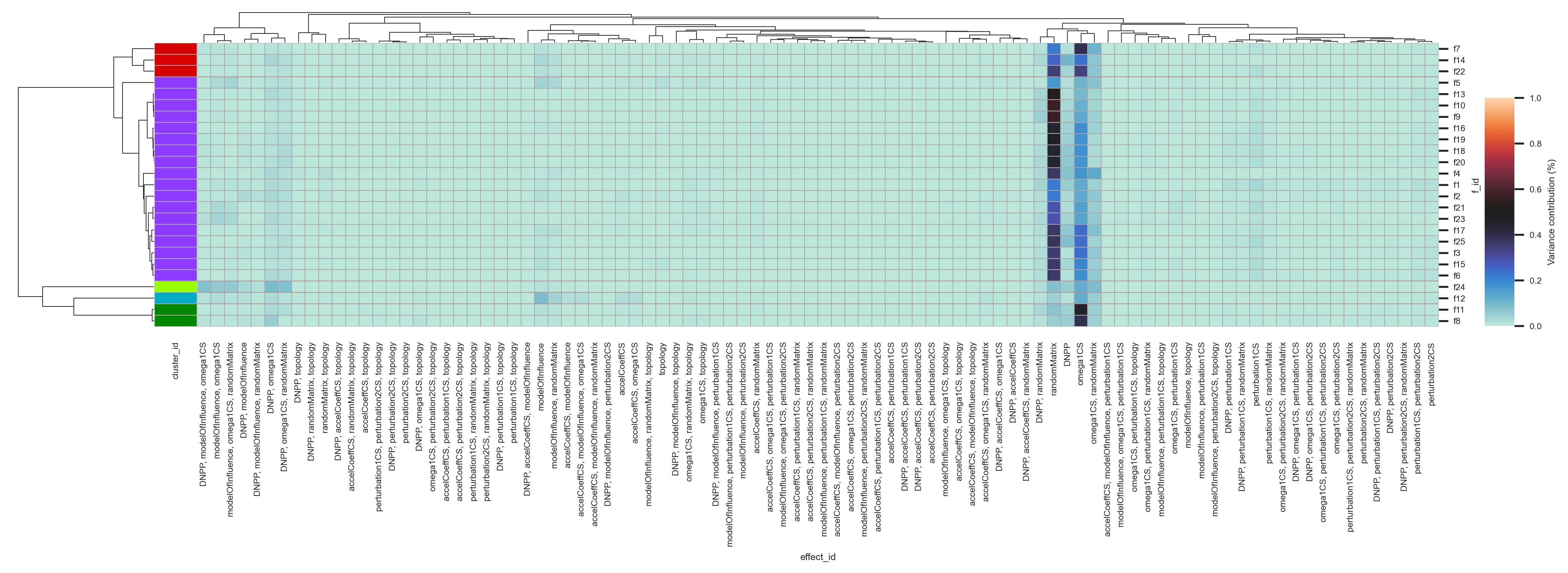}
    \caption{Clustermap of the 30$D$ problems.}
    \label{fig:clustermap30}
\end{figure*}

\subsubsection{Results on the 30$D$ Problems}\label{sec:Results on the 30D Problems}

\begin{figure*}[thb]
    \centering
    \makebox[\textwidth][c]{%
        {\includegraphics[width=0.15\textwidth]{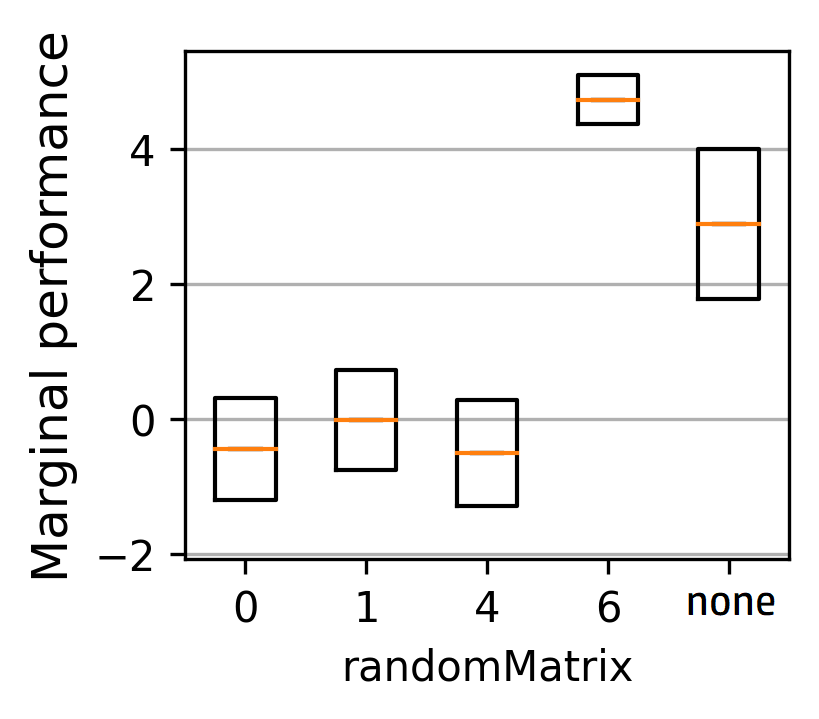}}\quad
        {\includegraphics[width=0.15\textwidth]{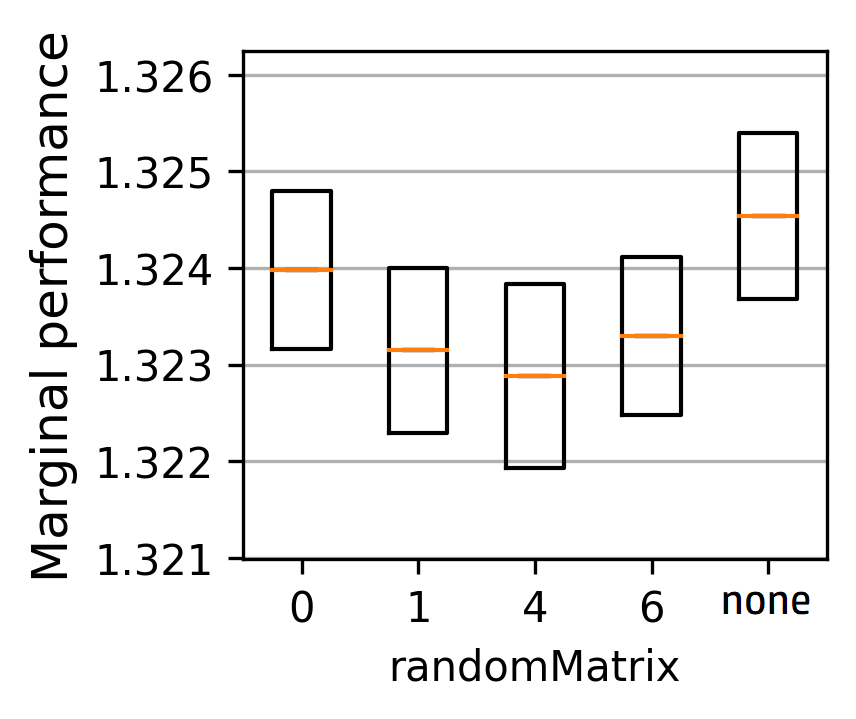}}\quad
        {\includegraphics[width=0.15\textwidth]{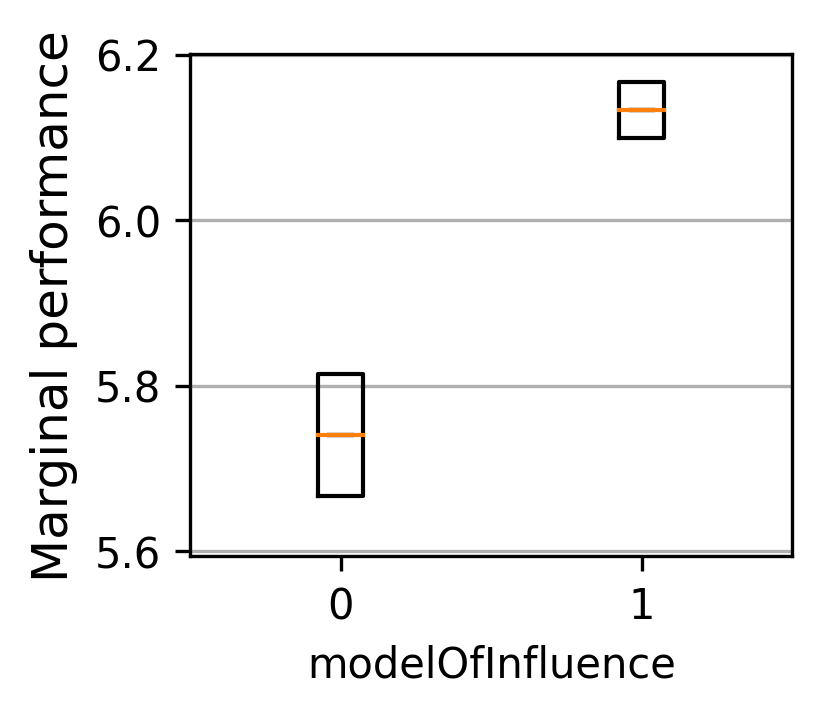}}\quad
        {\includegraphics[width=0.15\textwidth]{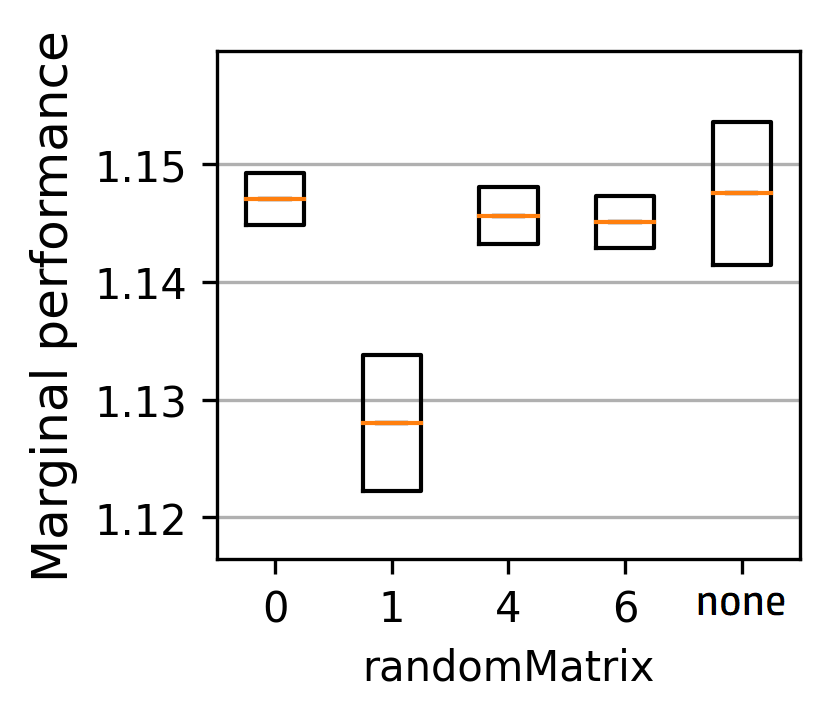}}\quad
        {\includegraphics[width=0.15\textwidth]{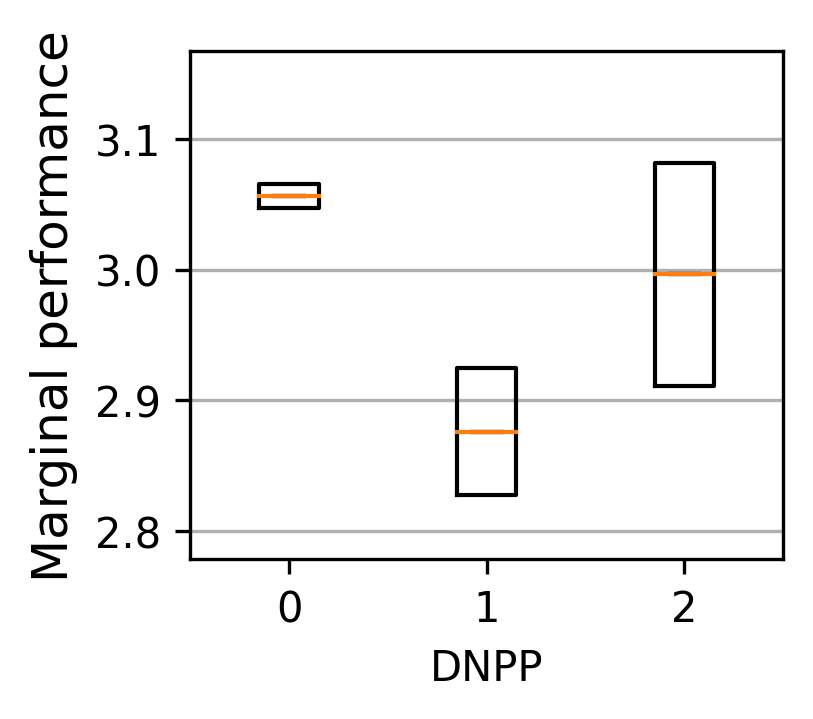}}
        }
    \makebox[\textwidth][c]{%
        {\includegraphics[width=0.15\textwidth]{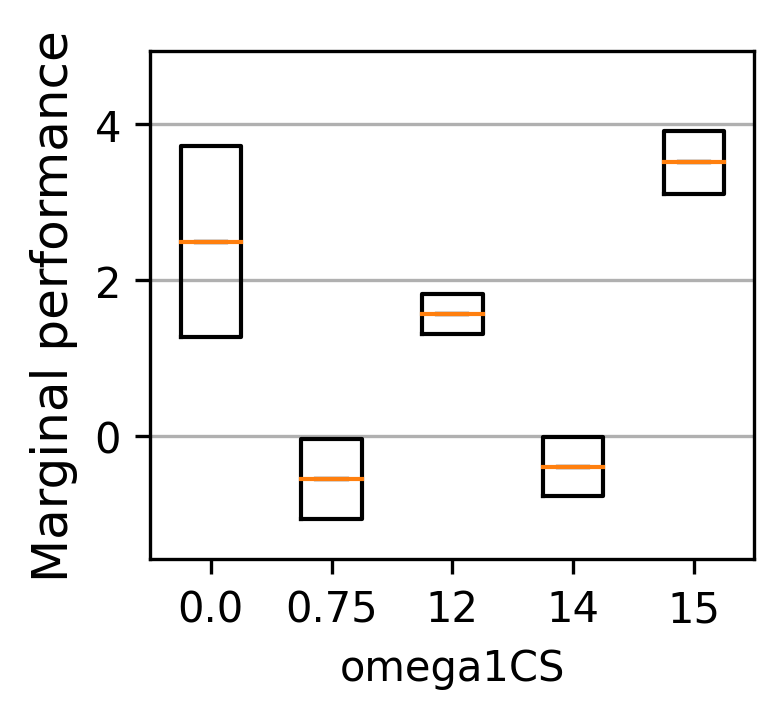}}\quad
        {\includegraphics[width=0.15\textwidth]{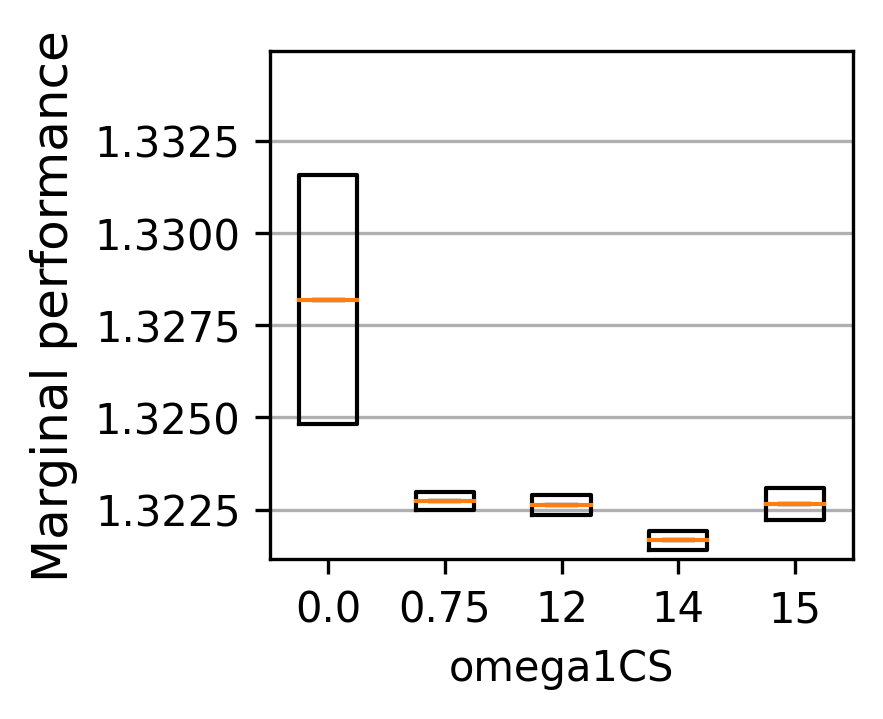}}\quad
        {\includegraphics[width=0.15\textwidth]{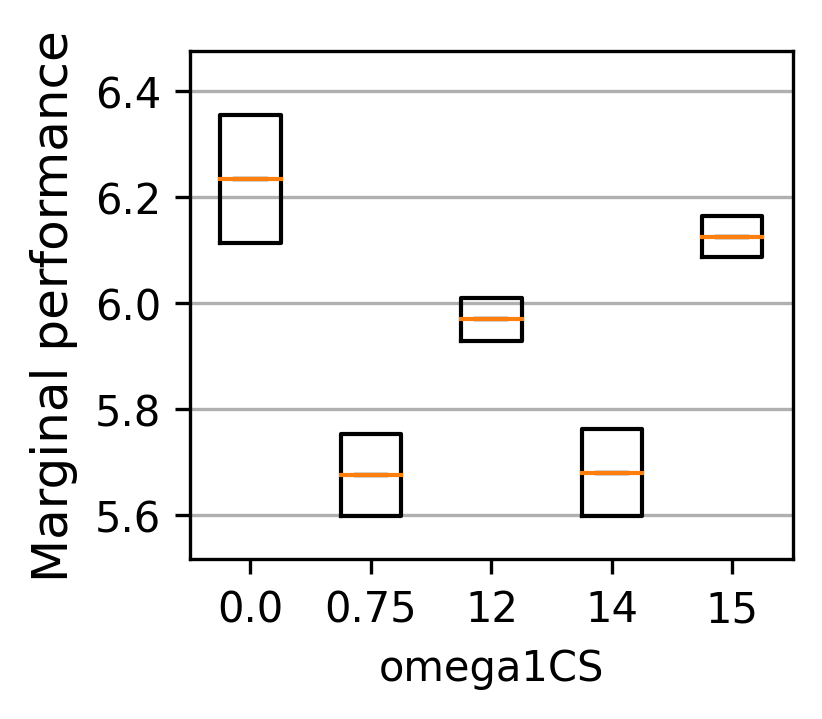}}\quad
        {\includegraphics[width=0.15\textwidth]{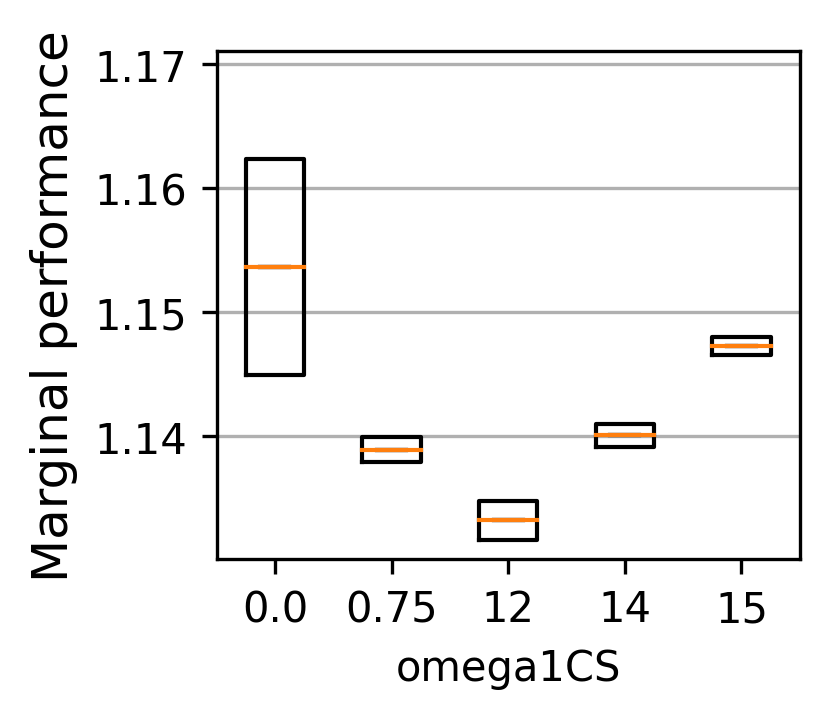}}\quad
        {\includegraphics[width=0.15\textwidth]{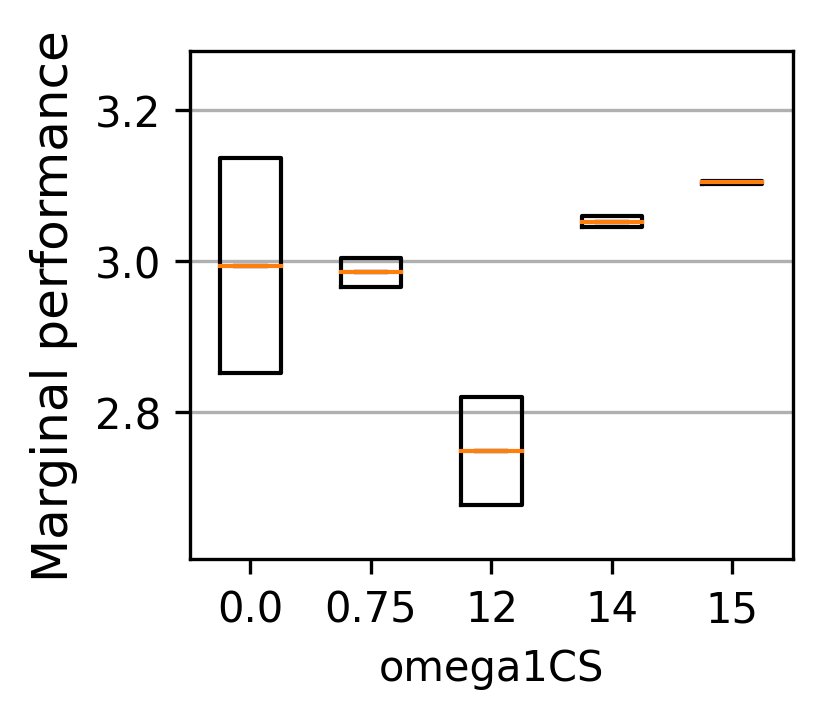}}
    }
    \makebox[\textwidth][c]{%
        \subfloat[$f_{1}$]
        {\includegraphics[width=0.17\textwidth]{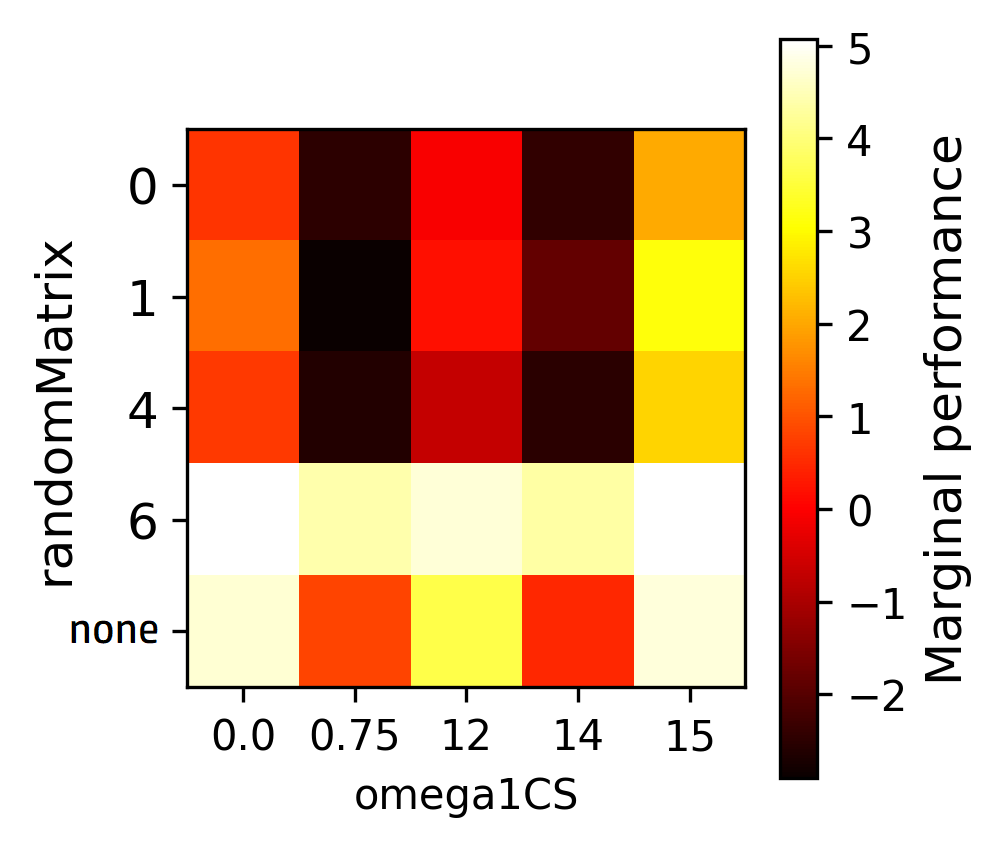}\label{fig:omega1CS_randomMatrix_0_30D}}\quad
        \subfloat[$f_{8}$]
        {\includegraphics[width=0.18\textwidth]{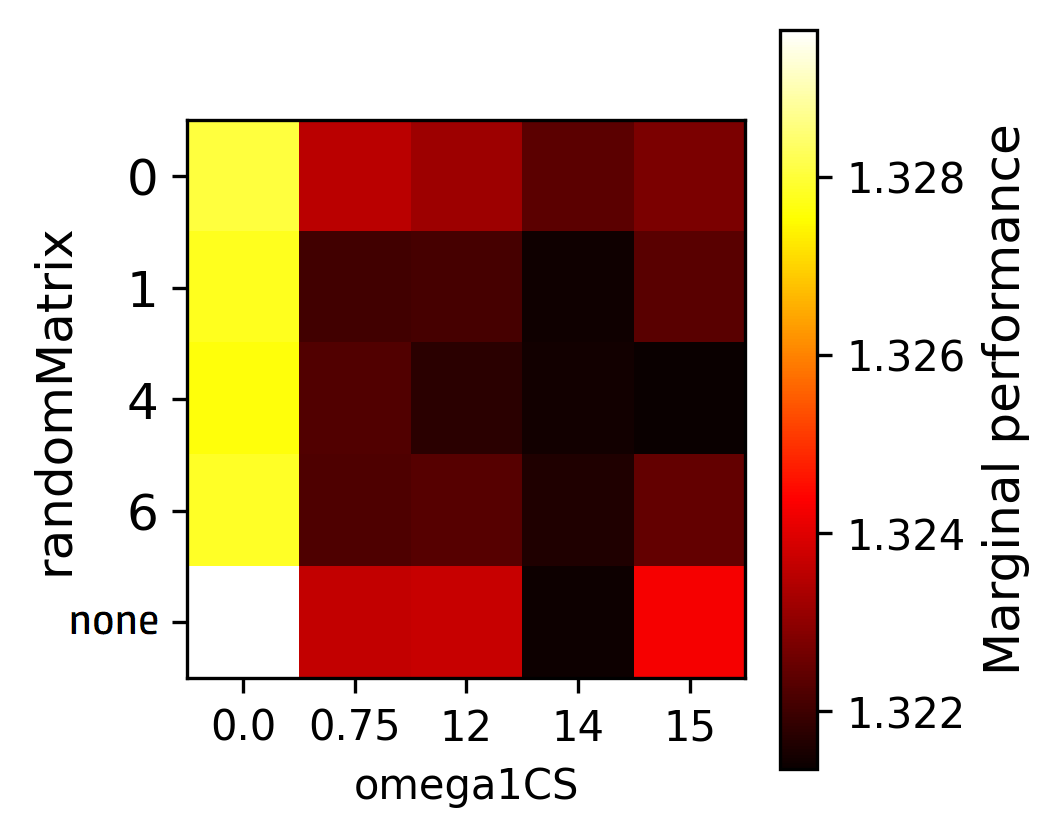}\label{fig:omega1CS_randomMatrix_7_30D}}\quad
        \subfloat[$f_{12}$]
        {\includegraphics[width=0.14\textwidth]{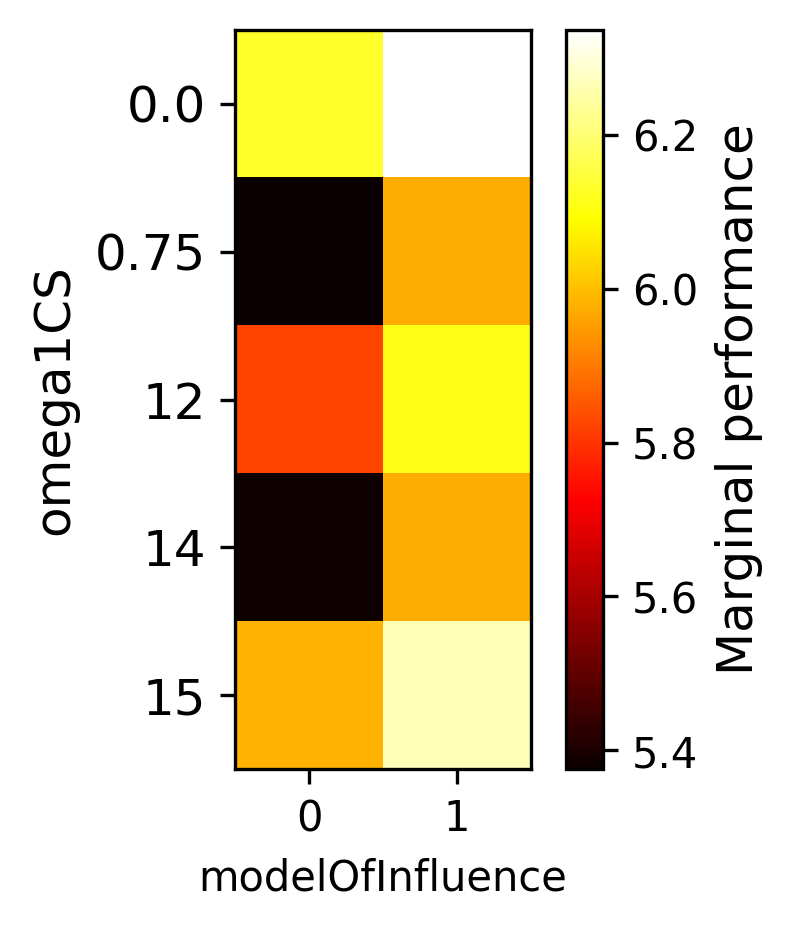}\label{fig:modelOfInfluence_omega1CS_11_30D}}\quad
        \subfloat[$f_{14}$]
        {\includegraphics[width=0.18\textwidth]{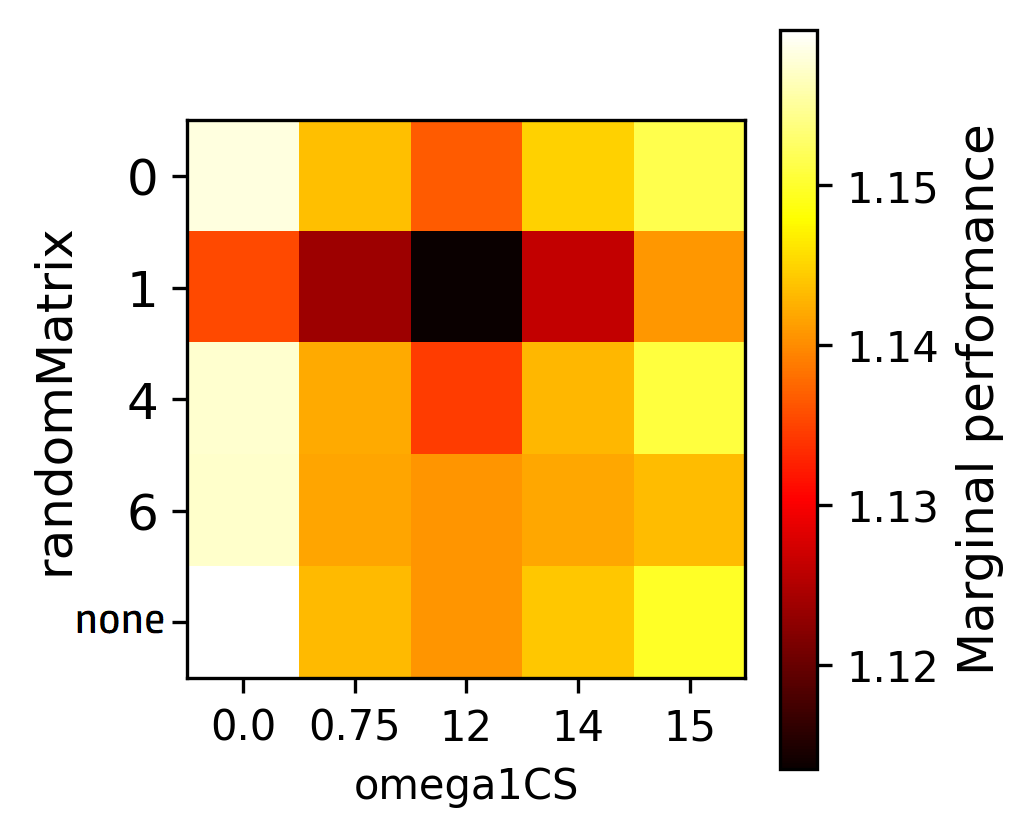}\label{fig:omega1CS_randomMatrix_13_30D}}\quad
        \subfloat[$f_{24}$]
        {\includegraphics[width=0.14\textwidth]{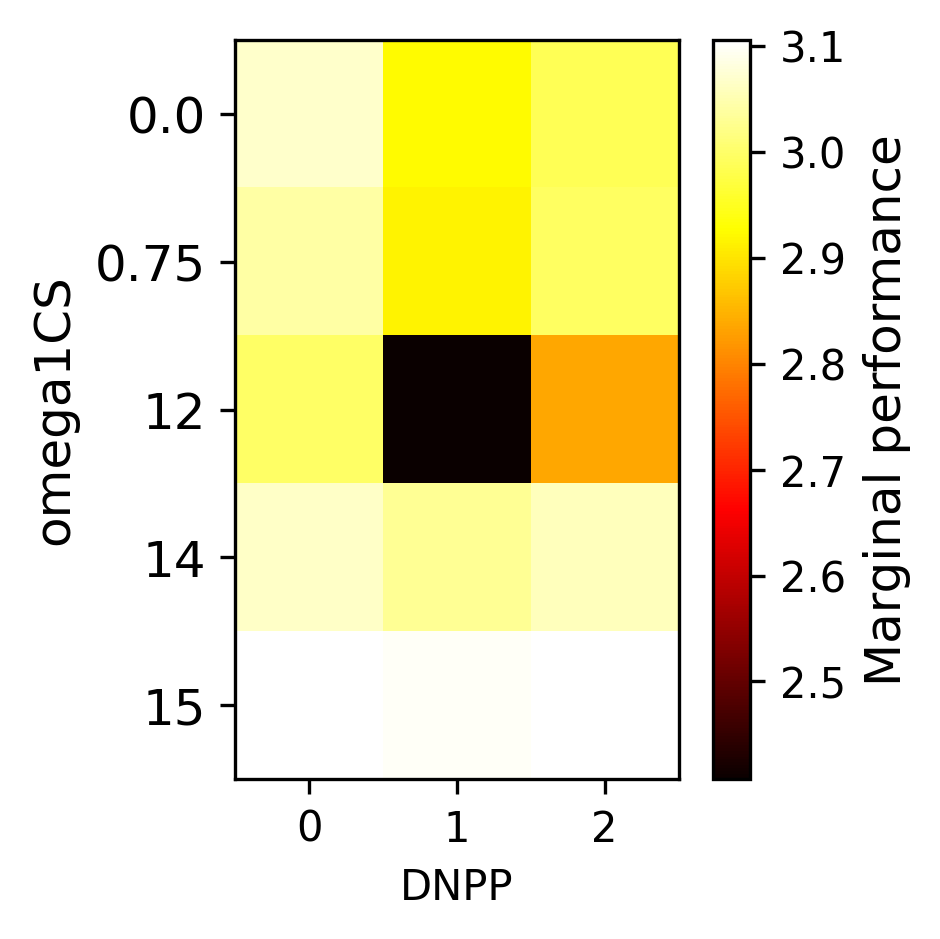}\label{fig:dnpp_omega1CS_23_30D}}
    }
    \caption{Marginal performance of the \rndMtx and \OmgCS modules in problem classes $f_1$ (\texttt{purple} cluster), $f_{8}$ (\texttt{dark green} cluster) and $f_{14}$ (\texttt{red} cluster); 
    the \Moi and \OmgCS modules in problem class $f_{12}$ (\texttt{cyan} cluster); 
    and the \DNPP and \OmgCS modules in problem class $f_{24}$ (\texttt{light green} cluster). All of them with $D=30$.}
    \label{fig:marginal_performance_30D}
\end{figure*}

In this section, we focus on the results obtained on the 30$D$ variants of the problems.
The goal is to identify the role that higher dimensionality plays in the activation of \PSOX modules and their marginal performance effects.
In the clustermap shown in \Fig\ref{fig:clustermap30}, we observe that there are three main clusters (\texttt{red}, \texttt{purple} and \texttt{dark green}) and two small clusters with only one function each (\texttt{light green} and \texttt{cyan}). 
Similarly to the 10$D$ results, variance decompositions confirm that the main effects of \OmgCS, \rndMtx, and \DNPP account for the majority of the performance differences.
In \Sect\ref{sec:Cumulative module importance}, we observed that the cumulative-variance for the 30$D$ problems resulted in curves that rise steeply within the first few effects and then saturate more gradually, indicating a modest, but noticeable, increase in modules activation due to the increase in dimensionality.
In \Fig\ref{fig:marginal_performance_30D}, to exemplify the individual and combined marginal performance, we selected the \rndMtx and \OmgCS modules in functions $f_1$ (\texttt{purple} cluster), $f_{8}$ (\texttt{dark green} cluster) and $f_{14}$ (\texttt{red} cluster), the \Moi and \OmgCS modules in function $f_{12}$ (\texttt{cyan} cluster), and the \DNPP and \OmgCS modules in function $f_{24}$ (\texttt{light green} cluster).

The implementation options of \OmgCS and \rndMtx that led to best performance on the 30$D$ problems are the same as those in the 10$D$ problems.
Also similarly to the results on the 10$D$ problems are the interaction effects that play the most impactful role in the 30$D$ problems, which involve combinations of \OmgCS $+$ \rndMtx, \DNPP $+$ \OmgCS, \DNPP $+$ \rndMtx, and the triplet \DNPP $+$ \OmgCS $+$ \rndMtx.
However, in the 30$D$ problems, we observe a slightly stronger activation of the \PertInfCS module and slightly weaker activation of the \Moi.
These two modules work together. 
The \PertInfCS module adds extra diversity to particle movement by computing a random vector centered around a neighboring solution, whereas the \Moi module determines the pool of neighboring solutions from which a particle can choose.
The stronger activation of \PertInfCS suggests that as the number of dimensions increases, particles benefit from higher levels of stochasticity in the social component, which makes the specific choice for the \Moi option less important.

In the case of the \DNPP module, the best options are $0=\DNPPRect$ and $1=\DNPPSphe$, with \DNPPRect showing higher individual effect in the unimodal functions $f_{1}$-$f_5$ and \DNPPSphe in the multimodal and hybrid functions. This trend is also present in the results on 10$D$, but it shows to be stronger in the 30$D$ problems.
It is worth noticing that, although the \DNPP determines the type of mapping used by the particles to sample new positions and whether \rndMtx and \PertInfCS are used in the implementations, the \DNPP module itself contributes very little to the performance of \PSOX.
This is because the \DNPP module serves as a template for other modules to interact with; thus, it indirectly affects \PSOX performance, while modules such as \rndMtx handle the primary workload.
 
In \Fig\ref{fig:clustermap30}, the \texttt{red} cluster includes functions $f_7$, $f_{14}$ and $f_{22}$, which exhibit similarly strong sensitivity to the \OmgCS the \rndMtx and their pairwise interaction. 
The three functions have multimodal, non-separable and rotated landscapes, while function $f_7$ is also unbounded and function $f_{22}$ ill-conditioned.
These characteristics makes all three functions particularly hard to solve.
In higher dimension, the attraction basins elongate and overlap more, making the role of \OmgCS crucial for particle to escape these basins and move to higher quality regions.
On the other hand, addressing multimodal and rotated landscapes requires selecting the right option for the \rndMtx module, since the complexity of estimating the rotation basis for their velocity vector ($\vec{v}^{i}_{t+1}$) increases together with the number of dimensions. 
The complicated interplay between \OmgCS and \rndMtx in this problems class can be visually appreciated in the pairwise interaction heatmap for function $f_{14}$, \Fig\ref{fig:omega1CS_randomMatrix_13_30D}, which shows that the higher marginal contribution effect is obtained only when combining options $1=\MtxDiagonal$ and $12=\OmegaAdapVel$.

The functions in the \texttt{dark green} cluster---$f_8$ and $f_{11}$---are multimodal with noisy landscapes and global optima placed in inaccessible locations.
Both functions exhibit elevated \OmgCS effects and modest, but similarly important contribution of the \rndMtx and \DNPP modules and their combined interaction.
For function $f_8$, there is also a noticeable effect of the interaction between the \DNPP and \OmgCS modules. The results for these two functions on the 30$D$ problems are similar to the ones on the 10$D$, highlighting the fact that dimensionality is not a defining feature in the complexity of functions $f_8$ and $f_{11}$.
In this problem class, particles' ability to navigate noisy and highly rugged structure by controlling their inertia is the biggest performance driver.
This can be observed in the boxplots and heatmaps for function $f_8$, \Fig\ref{fig:omega1CS_randomMatrix_7_30D}, which show that the use of option $12=\OmegaAdapVel$ is the main one influencing the performance of \PSOX.

The largest cluster in \Fig\ref{fig:clustermap30} is the \texttt{purple} cluster, which comprises functions $f_{1}$-$f_{6}$, $f_{9}$, $f_{10}$, $f_{13}$, $f_{15}$-$f_{21}$,$f_{23}$ and $f_{25}$.
In this cluster, the most impactful module is \rndMtx, which accounts for 50-60\% of the variance in performance in 13 out of the 18 function in the cluster.
Although the individual effect of \rndMtx is evident in the vast majority of multimodal and hybrid function, we observe that its effect diminishes (less than 30\% of the variance) in three unimodal functions ($f_{1}$, $f_{2}$, $f_{5}$) and two hybrid compositions ($f_{21}$ and $f_{23}$).
In fact, in the subset of functions composed of $f_{1}$, $f_{2}$, $f_{5}$, $f_{21}$ and $f_{23}$, the individual effect of \OmgCS, the pairwise $\OmgCS+\rndMtx$ and $\DNPP+\OmgCS$ effects, and triple $\DNPP+\OmgCS+\rndMtx$ effect are the most impactful.
The \PertInfCS module is most active in the \texttt{purple} cluster compared to 
the rest of the clusters, however, its contribution is incremental at most.

Function $f_{12}$, in the \texttt{cyan} cluster, and functions $f_{24}$, in the \texttt{light green} cluster remain outlier in 30$D$.
They show a similar activation of modules that in the 10$D$ results, but with a stronger marginal effect.
Function $f_{12}$ is the only one in which we can observe the influence of other modules to contribute to performance in similar amount as \rndMtx or \OmgCS.
In function $f_{12}$, the \Moi module, its combined effect with \rndMtx and the \AcCof module are also predominant, while the \DNPP module shows to have no impact whatsoever.
In the case of $f_{24}$, the pairwise and triple interactions of \rndMtx, \OmgCS, and \DNPP show similar variance contribution as their individual effect.
This suggests that function $f_{24}$ may be particularly sensitive to the implementation design, since it involves three modules working in combination.

\section{Conclusions and Outlook}\label{sec:Conclusions}
We quantified how algorithmic modules in the \PSOX framework contribute to performance across the CEC’05 problem classes. Using \fANV over a large design space of 1{,}424 automatically generated \PSOX variants, we decomposed performance variability into isolated, pairwise, and triple interaction effects of key modules. Across both 10$D$ and 30$D$ scenarios, our cumulative variance analysis showed that a small set of module effects (namely the \rndMtx, \OmgCS and \DNPP modules) accounts for most of the explainable variance, with single effects dominating the early contribution and pairwise interactions adding explanatory power thereafter. Triple interactions were present but generally weak, indicating that most of the variance in performance is captured by the main and pairwise interactions.

The clustering analysis of per-function importance profiles revealed groups of problem classes that share similar module-effect patterns, even when these groups only partially align with the CEC'05 high-level features categorization. The \rndMtx modules exhibited the higher marginal effect on most functions, followed by the \OmgCS module and finally the \DNPP module. Our results provide two forms of guidance on \PSO. 
From a design standpoint, selecting and configuring the \rndMtx and \OmgCS modules is critical when instantiating \PSOX variants, while the \DNPP module can be fixed based on the anticipated complexity---\DNPPRect if it is \textit{easy} and \DNPPSphe if it is \textit{hard}. 
From a selection standpoint, the observed clusters suggest that module choices transfer within groups of functions that share similar importance patterns, even when those groups do not coincide perfectly with classical property-based partitions. In practice, this means that one can use a portfolio strategy to implement \PSO, in which a set of \PSOX configurations---diversified primarily along \rndMtx and \OmgCS options and secondarily along the \DNPP and \Moi options---can cover most problem classes with minimal redundancy.

The main limitation of our study is that it is tied to the CEC’05 suite and to the specific \PSOX modules considered. Clearly, different problem families, extended module spaces and larger parameter configuration ranges may reshape the importance landscape. Moreover, while triple interactions were measured, higher-order effects remain uncharted and they may be present in very different experimental conditions. 
In future research, we will: (i) enlarge the \PSOX module set to test additional modules, with a focus on rotations and population controls schemes; (ii) replicate the analysis on other benchmark families, including BBOB and CEC suites, and on other modular framework, such as \metafor; and (iii) tighten the link between problem and algorithm by creating enhanced meta-features that enable principled, data-driven module selection.




\begin{thebibliography}{43}
\providecommand{\natexlab}[1]{#1}
\providecommand{\url}[1]{#1}
\csname url@samestyle\endcsname
\providecommand{\newblock}{\relax}
\providecommand{\bibinfo}[2]{#2}
\providecommand{\BIBentrySTDinterwordspacing}{\spaceskip=0pt\relax}
\providecommand{\BIBentryALTinterwordstretchfactor}{4}
\providecommand{\BIBentryALTinterwordspacing}{\spaceskip=\fontdimen2\font plus
\BIBentryALTinterwordstretchfactor\fontdimen3\font minus \fontdimen4\font\relax}
\providecommand{\BIBforeignlanguage}[2]{{%
\expandafter\ifx\csname l@#1\endcsname\relax
\typeout{** WARNING: IEEEtranN.bst: No hyphenation pattern has been}%
\typeout{** loaded for the language `#1'. Using the pattern for}%
\typeout{** the default language instead.}%
\else
\language=\csname l@#1\endcsname
\fi
#2}}
\providecommand{\BIBdecl}{\relax}
\BIBdecl

\bibitem[Luenberger et~al.(2016)Luenberger, Ye, et~al.]{LueYe1984:Book-linear-nonlinear}
D.~G. Luenberger, Y.~Ye \emph{et~al.}, \emph{Linear and Nonlinear Programming}.\hskip 1em plus 0.5em minus 0.4em\relax Cham: Springer, 2016.

\bibitem[Audet and Hare(2017)]{AudHar2017:Book-BBO-DFO}
C.~Audet and W.~Hare, \emph{Derivative-free and blackbox optimization}.\hskip 1em plus 0.5em minus 0.4em\relax Berlin, Heidelberg, Germany: Springer, 2017.

\bibitem[Rechenberg(1971)]{Rec1971PhD}
I.~Rechenberg, ``Evolutionsstrategie: {O}ptimierung technischer {S}ysteme nach {P}rinzipien der biologischen {E}volution,'' Ph.D. dissertation, Department of Process Engineering, Technical University of Berlin, 1971.

\bibitem[Schwefel(1977)]{Schwefel1977}
H.-P. Schwefel, \emph{Numerische {O}ptimierung von {C}omputer--{M}odellen mittels der {E}volutionsstrategie}.\hskip 1em plus 0.5em minus 0.4em\relax Basel, Switzerland: Birkh{\"a}user, 1977.

\bibitem[Storn and Price(1997)]{StoPri1997:de}
R.~Storn and K.~Price, ``Differential evolution -- a simple and efficient heuristic for global optimization over continuous spaces,'' \emph{Journal of Global Optimization}, vol.~11, no.~4, pp. 341--359, 1997.

\bibitem[Kennedy and Eberhart(1995)]{KenEbe1995pso}
J.~Kennedy and R.~Eberhart, ``Particle swarm optimization,'' in \emph{Proceedings of ICNN'95-International Conference on Neural Networks}, vol.~4.\hskip 1em plus 0.5em minus 0.4em\relax Piscataway, NJ: IEEE, 1995, pp. 1942--1948.

\bibitem[Eberhart and Kennedy(1995)]{EbeKen1995:pso}
R.~Eberhart and J.~Kennedy, ``A new optimizer using particle swarm theory,'' in \emph{Proceedings of the Sixth International Symposium on Micro Machine and Human Science}.\hskip 1em plus 0.5em minus 0.4em\relax Piscataway, NJ: IEEE Press, 1995, pp. 39--43.

\bibitem[Birattari et~al.(2002)Birattari, St{\"u}tzle, Paquete, and Varrentrapp]{BirStuPaqVar02:gecco}
M.~Birattari, T.~St{\"u}tzle, L.~Paquete, and K.~Varrentrapp, ``A racing algorithm for configuring metaheuristics,'' in \emph{Proceedings of the Genetic and Evolutionary Computation Conference, GECCO 2002}, W.~B. Langdon \emph{et~al.}, Eds.\hskip 1em plus 0.5em minus 0.4em\relax Morgan Kaufmann Publishers, San Francisco, CA, 2002, pp. 11--18.

\bibitem[Birattari(2009)]{Birattari09tuning}
M.~Birattari, \emph{Tuning Metaheuristics: A Machine Learning Perspective}, ser. Studies in Computational Intelligence.\hskip 1em plus 0.5em minus 0.4em\relax Berlin\slash Heidelberg, Germany: Springer, 2009, vol. 197.

\bibitem[L{\'o}pez-Ib{\'a}{\~n}ez and St{\"u}tzle(2010)]{LopStu2010:gecco}
M.~L{\'o}pez-Ib{\'a}{\~n}ez and T.~St{\"u}tzle, ``The impact of design choices of multi-objective ant colony optimization algorithms on performance: An experimental study on the biobjective {TSP},'' in \emph{Proceedings of the Genetic and Evolutionary Computation Conference, GECCO 2010}, M.~Pelikan and J.~Branke, Eds.\hskip 1em plus 0.5em minus 0.4em\relax New York, NY: ACM Press, 2010, pp. 71--78.

\bibitem[Hooker(1996)]{Hoo1996joh}
J.~N. Hooker, ``Testing heuristics: We have it all wrong,'' \emph{Journal of Heuristics}, vol.~1, no.~1, pp. 33--42, 1996.

\bibitem[Garc{\'i}a-Mart{\'i}nez et~al.(2017)Garc{\'i}a-Mart{\'i}nez, Guti{\'e}rrez, Molina, Lozano, and Herrera]{GarGutMol2017}
C.~Garc{\'i}a-Mart{\'i}nez, P.~D. Guti{\'e}rrez, D.~Molina, M.~Lozano, and F.~Herrera, ``Since {CEC} 2005 competition on real-parameter optimisation: a decade of research, progress and comparative analysis's weakness,'' \emph{Soft Computing}, vol.~21, no.~19, pp. 5573--5583, 2017.

\bibitem[Nikolikj et~al.(2025)Nikolikj, Munoz, and Eftimov]{Nikolikj2025:benchmarking}
A.~Nikolikj, M.~A. Munoz, and T.~Eftimov, ``Benchmarking footprints of continuous black-box optimization algorithms: Explainable insights into algorithm success and failure,'' \emph{Swarm and Evolutionary Computation}, vol.~94, p. 101895, 2025.

\bibitem[Doerr et~al.(2018)Doerr, Wang, Ye, Van~Rijn, and B{\"a}ck]{DoeWanYe2018:IOHprofiler}
C.~Doerr, H.~Wang, F.~Ye, S.~Van~Rijn, and T.~B{\"a}ck, ``{IOH}profiler: A benchmarking and profiling tool for iterative optimization heuristics,'' \emph{arXiv preprint arXiv:1810.05281}, 2018.

\bibitem[Bartz-Beielstein et~al.(2020)Bartz-Beielstein, Doerr, Berg, Bossek, Chandrasekaran, Eftimov, Fischbach, Kerschke, La~Cava, Lopez-Ibanez, et~al.]{BarDoeBer2020:BenchmarkingInOptimization}
T.~Bartz-Beielstein, C.~Doerr, D.~v.~d. Berg, J.~Bossek, S.~Chandrasekaran, T.~Eftimov, A.~Fischbach, P.~Kerschke, W.~La~Cava, M.~Lopez-Ibanez \emph{et~al.}, ``Benchmarking in optimization: Best practice and open issues,'' \emph{arXiv preprint arXiv:2007.03488}, 2020.

\bibitem[Camacho-Villalón et~al.(2023)Camacho-Villalón, Stützle, and Dorigo]{CamStuDor2023:IC:disNewMH}
C.~L. Camacho-Villalón, T.~Stützle, and M.~Dorigo, ``Designing new metaheuristics: Manual versus automatic approaches,'' \emph{Intelligent Computing}, vol.~2, no. 0048, pp. 1--15, 2023.

\bibitem[{Camacho-Villal{\'o}n} et~al.(2022){Camacho-Villal{\'o}n}, Dorigo, and St{\"u}tzle]{CamDorStu2022:tec}
C.~L. {Camacho-Villal{\'o}n}, M.~Dorigo, and T.~St{\"u}tzle, ``{PSO}\textsl{-X}: A component-based framework for the automatic design of particle swarm optimization algorithms,'' \emph{IEEE Transactions on Evolutionary Computation}, vol.~26, no.~3, pp. 402--416, 2022.

\bibitem[Hutter et~al.(2014)Hutter, Hoos, and Leyton-Brown]{HutHooLey2014icml}
\BIBentryALTinterwordspacing
F.~Hutter, H.~H. Hoos, and K.~Leyton-Brown, ``An efficient approach for assessing hyperparameter importance,'' in \emph{Proceedings of the 31th International Conference on Machine Learning}, vol.~32, 2014, pp. 754--762. [Online]. Available: \url{http://jmlr.org/proceedings/papers/v32/hutter14.html}
\BIBentrySTDinterwordspacing

\bibitem[Van~Rijn and Hutter(2018)]{RijHut18:sigkdd}
J.~N. Van~Rijn and F.~Hutter, ``Hyperparameter importance across datasets,'' in \emph{Proceedings of the 24th ACM SIGKDD international conference on knowledge discovery \& data mining}, 2018, pp. 2367--2376.

\bibitem[Suganthan et~al.(2005)Suganthan, Hansen, Liang, Deb, Chen, Auger, and Tiwari]{SugHanLia2005cec}
P.~N. Suganthan, N.~Hansen, J.~J. Liang, K.~Deb, Y.~P. Chen, A.~Auger, and S.~Tiwari, ``Problem definitions and evaluation criteria for the {CEC 2005} special session on real-parameter optimization,'' Nanyang Technological University, Singapore, Tech. Rep., 2005.

\bibitem[Banks et~al.(2007)Banks, Vincent, and Anyakoha]{BanVinAnY2007:natc-part1}
A.~Banks, J.~Vincent, and C.~Anyakoha, ``A review of particle swarm optimization. part i: background and development,'' \emph{Natural Computing}, vol.~6, no.~4, pp. 467--484, 2007.

\bibitem[Banks et~al.(2008)Banks, Vincent, and Anyakoha]{BanVinAnY2008:natc-part2}
------, ``A review of particle swarm optimization. part ii: hybridisation, combinatorial, multicriteria and constrained optimization, and indicative applications,'' \emph{Natural Computing}, vol.~7, no.~1, pp. 109--124, 2008.

\bibitem[Bonyadi and Michalewicz(2017)]{BonMic2017:mit}
M.~R. Bonyadi and Z.~Michalewicz, ``Particle swarm optimization for single objective continuous space problems: a review,'' 2017.

\bibitem[Engelbrecht(2013)]{Eng2013:brics}
A.~P. Engelbrecht, ``Particle swarm optimization: Global best or local best?'' in \emph{2013 BRICS congress on computational intelligence and 11th Brazilian congress on computational intelligence}, 2013, pp. 124--135.

\bibitem[St{\"u}tzle and L{\'o}pez-Ib{\'a}{\~n}ez(2019)]{StuLop2019hb}
T.~St{\"u}tzle and M.~L{\'o}pez-Ib{\'a}{\~n}ez, ``Automated design of metaheuristic algorithms,'' in \emph{Handbook of Metaheuristics}, ser. International Series in Operations Research \& Management Science, M.~Gendreau and J.-Y. Potvin, Eds.\hskip 1em plus 0.5em minus 0.4em\relax New York, NY: Springer, 2019, vol. 272, pp. 541--579.

\bibitem[de~Nobel et~al.(2021)de~Nobel, Vermetten, Wang, Doerr, and B{\"a}ck]{deNVerWan2021:gecco-CMAESframework}
J.~de~Nobel, D.~Vermetten, H.~Wang, C.~Doerr, and T.~B{\"a}ck, ``Tuning as a means of assessing the benefits of new ideas in interplay with existing algorithmic modules,'' in \emph{GECCO'21 Companion}, F.~Chicano, Ed.\hskip 1em plus 0.5em minus 0.4em\relax New York, NY: ACM Press, 2021, pp. 1375--1384.

\bibitem[Vermetten et~al.(2023)Vermetten, Caraffini, Kononova, and B{\"a}ck]{VerCarKon2023:autoDE:gecco}
D.~Vermetten, F.~Caraffini, A.~V. Kononova, and T.~B{\"a}ck, ``Modular differential evolution,'' in \emph{Proceedings of the Genetic and Evolutionary Computation Conference}.\hskip 1em plus 0.5em minus 0.4em\relax New York, NY: ACM Press, 2023, pp. 864--872.

\bibitem[Camacho-Villal{\'o}n et~al.(2025)Camacho-Villal{\'o}n, Dorigo, and St{\"u}tzle]{CamDorStu:25:telo}
C.~Camacho-Villal{\'o}n, M.~Dorigo, and T.~St{\"u}tzle, ``Metafor: A hybrid metaheuristics software framework for single-objective continuous optimization problems,'' \emph{arXiv preprint arXiv:2502.11225}, 2025.

\bibitem[L{\'o}pez-Ib{\'a}{\~n}ez et~al.(2016)L{\'o}pez-Ib{\'a}{\~n}ez, Dubois-Lacoste, {P{\'e}rez C{\'a}ceres}, St{\"u}tzle, and Birattari]{LopDubPerStuBir2016irace}
M.~L{\'o}pez-Ib{\'a}{\~n}ez, J.~Dubois-Lacoste, L.~{P{\'e}rez C{\'a}ceres}, T.~St{\"u}tzle, and M.~Birattari, ``The {\Rpackage{irace}} package: Iterated racing for automatic algorithm configuration,'' \emph{Operations Research Perspectives}, vol.~3, pp. 43--58, 2016.

\bibitem[Hutter et~al.(2009)Hutter, Hoos, Leyton-Brown, and St{\"u}tzle]{HutHooLeyStu2009jair}
F.~Hutter, H.~H. Hoos, K.~Leyton-Brown, and T.~St{\"u}tzle, ``{ParamILS:} an automatic algorithm configuration framework,'' \emph{Journal of Artificial Intelligence Research}, vol.~36, pp. 267--306, Oct. 2009.

\bibitem[Hutter et~al.(2011)Hutter, Hoos, and Leyton-Brown]{HutHooLey2011lion}
F.~Hutter, H.~H. Hoos, and K.~Leyton-Brown, ``Sequential model-based optimization for general algorithm configuration,'' in \emph{Learning and Intelligent Optimization, 5th International Conference, LION 5}, ser. Lecture Notes in Computer Science, C.~A. {Coello Coello}, Ed.\hskip 1em plus 0.5em minus 0.4em\relax Heidelberg, Germany: Springer, Heidelberg, Germany, 2011, vol. 6683, pp. 507--523.

\bibitem[{Camacho-Villal{\'o}n} et~al.(2021){Camacho-Villal{\'o}n}, Dorigo, and St{\"u}tzle]{CamDorStu2021:posx-supp}
C.~L. {Camacho-Villal{\'o}n}, M.~Dorigo, and T.~St{\"u}tzle, ``Pso\textsl{-X}: A component-based framework for the automatic design of particle swarm optimization algorithms: Supplementary material,'' \url{http://iridia.ulb.ac.be/supp/IridiaSupp2021-001/}, 2021.

\bibitem[Hutter et~al.(2013)Hutter, Hoos, and Leyton-Brown]{HutHooLey2013lion}
F.~Hutter, H.~H. Hoos, and K.~Leyton-Brown, ``Identifying key algorithm parameters and instance features using forward selection,'' in \emph{Learning and Intelligent Optimization, 7th International Conference, LION 7}, ser. Lecture Notes in Computer Science, P.~M. Pardalos and G.~Nicosia, Eds., vol. 7997.\hskip 1em plus 0.5em minus 0.4em\relax Springer, Heidelberg, Germany, 2013, pp. 364--381.

\bibitem[Siegmund et~al.(2015)Siegmund, Grebhahn, Apel, and K{\"a}stner]{Siegmund2015}
N.~Siegmund, A.~Grebhahn, S.~Apel, and C.~K{\"a}stner, ``Performance-influence models for highly configurable systems,'' in \emph{Proceedings of the 2015 10th Joint Meeting on Foundations of Software Engineering}, 2015, pp. 284--294.

\bibitem[Biedenkapp et~al.(2017)Biedenkapp, Lindauer, Eggensperger, Hutter, Fawcett, and Hoos]{BieLinEggFraFawHoo2017}
\BIBentryALTinterwordspacing
A.~Biedenkapp, M.~Lindauer, K.~Eggensperger, F.~Hutter, C.~Fawcett, and H.~H. Hoos, ``Efficient parameter importance analysis via ablation with surrogates,'' in \emph{AAAI Conference on Artificial Intelligence}, S.~P. Singh and S.~Markovitch, Eds.\hskip 1em plus 0.5em minus 0.4em\relax {AAAI} Press, Feb. 2017. [Online]. Available: \url{https://aaai.org/ocs/index.php/AAAI/AAAI17/paper/view/14750}
\BIBentrySTDinterwordspacing

\bibitem[Prager et~al.(2020)Prager, Trautmann, Wang, B{\"a}ck, and Kerschke]{Prager2020}
R.~P. Prager, H.~Trautmann, H.~Wang, T.~H. B{\"a}ck, and P.~Kerschke, ``Per-instance configuration of the modularized cma-es by means of classifier chains and exploratory landscape analysis,'' in \emph{2020 IEEE Symposium Series on Computational Intelligence (SSCI)}.\hskip 1em plus 0.5em minus 0.4em\relax IEEE, 2020, pp. 996--1003.

\bibitem[Kostovska et~al.(2022)Kostovska, Vermetten, D{\v{z}}eroski, Doerr, Korosec, and Eftimov]{KosVerDze22:gecco}
A.~Kostovska, D.~Vermetten, S.~D{\v{z}}eroski, C.~Doerr, P.~Korosec, and T.~Eftimov, ``The importance of landscape features for performance prediction of modular cma-es variants,'' in \emph{Proceedings of the Genetic and Evolutionary Computation Conference}, 2022, pp. 648--656.

\bibitem[Kostovska et~al.(2025)Kostovska, Doerr, D\v{z}eroski, Panov, and Eftimov]{Kostovksa2025}
A.~Kostovska, C.~Doerr, S.~D\v{z}eroski, P.~Panov, and T.~Eftimov, ``Geometric learning in black-box optimization: A gnn framework for algorithm performance prediction,'' in \emph{Proceedings of the Genetic and Evolutionary Computation Conference Companion}, 2025, p. In Press.

\bibitem[Van~Stein et~al.(2024)Van~Stein, Vermetten, V.~Kononova, and B{\"a}ck]{SteVerKon24:xai-heu}
N.~Van~Stein, D.~Vermetten, A.~V.~Kononova, and T.~B{\"a}ck, ``Explainable benchmarking for iterative optimization heuristics,'' \emph{ACM Transactions on Evolutionary Learning}, 2024.

\bibitem[Nikolikj et~al.(2024)Nikolikj, Kostovska, Vermetten, Doerr, and Eftimov]{NikKosVer24:cec}
A.~Nikolikj, A.~Kostovska, D.~Vermetten, C.~Doerr, and T.~Eftimov, ``Quantifying individual and joint module impact in modular optimization frameworks,'' in \emph{2024 IEEE Congress on Evolutionary Computation (CEC)}.\hskip 1em plus 0.5em minus 0.4em\relax IEEE, 2024, pp. 1--8.

\bibitem[Nikolikj and Eftimov(2025)]{Nikolikj2025:sec}
A.~Nikolikj and T.~Eftimov, ``Exploring module interactions in modular cma-es across problem classes,'' \emph{Swarm and Evolutionary Computation}, vol.~98, p. 102116, 2025.

\bibitem[M{\"u}llner(2011)]{Mul11:hc}
D.~M{\"u}llner, ``Modern hierarchical, agglomerative clustering algorithms,'' \emph{arXiv preprint arXiv:1109.2378}, 2011.

\bibitem[Rousseeuw(1987)]{Rou87:silhouettes}
P.~J. Rousseeuw, ``Silhouettes: a graphical aid to the interpretation and validation of cluster analysis,'' \emph{Journal of computational and applied mathematics}, vol.~20, pp. 53--65, 1987.

\end{thebibliography}

\providecommand{\MaxMinAntSystem}{{$\cal MAX$--$\cal MIN$} {A}nt {S}ystem} \providecommand{\Rpackage}[1]{#1} \providecommand{\SoftwarePackage}[1]{#1} \providecommand{\proglang}[1]{#1}

\end{document}